\documentclass{article}

\usepackage{jmlr2e}

\usepackage[utf8]{inputenc} 
\usepackage[T1]{fontenc}    
\usepackage[colorlinks,citecolor=blue,urlcolor=blue,pagebackref]{hyperref} 
\hypersetup{
  colorlinks,
  linkcolor={blue!50!black},
  citecolor={blue!50!black},
  urlcolor={blue!80!black}
}

\renewcommand*{\backref}[1]{\ifx#1\relax \else Page #1 \fi}
\renewcommand*{\backrefalt}[4]{
  \ifcase #1 \footnotesize{(Not cited.)}
  \or        \footnotesize{(Cited on page~#2.)}
  \else      \footnotesize{(Cited on pages~#2.)}
  \fi
}

\usepackage{url}      
\usepackage{booktabs} 
\usepackage{amsfonts} 
\usepackage{nicefrac} 
\usepackage{microtype}
\usepackage{xcolor}   

\usepackage[linesnumbered,ruled,vlined]{algorithm2e}
\SetKwInput{KwInput}{Input}
\usepackage{algpseudocode}

\usepackage{amsmath,amsthm,amsfonts,amssymb,dsfont,bm}
\usepackage{enumerate,enumitem,color}
\usepackage{xcolor}
\usepackage{todonotes}

\usepackage{graphicx}
\usepackage{subfigure,wrapfig}

\usepackage{cleveref}

\usepackage[normalem]{ulem}

\usepackage[skins,breakable]{tcolorbox}

\usepackage{silence}
\ActivateWarningFilters[warnfilt]

\usepackage{todonotes}
\newcommand{\mg}[1]{\textcolor{black}{#1}}

\newcommand{\dimh}{\dim_{\mathrm{H}}}
\newcommand{\dimhu}{\overline{\dim}_{\mathrm{H}}}
\newcommand{\dimhl}{\underline{\dim}_{\mathrm{H}}}
\newcommand{\dimm}{\dim_{\mathrm{M}}}
\newcommand{\dimmu}{\overline{\dim}_{\mathrm{M}}}

\newcommand{\dimlu}{\overline{\dim}_{\mathrm{loc}}}
\newcommand{\dimll}{\underline{\dim}_{\mathrm{loc}}}
\newcommand{\diml}{\dim_{\mathrm{loc}}}

\newcommand{\wb}{w}

\newcommand{\rset}{\mathbb{R}}
\newcommand{\pr}{\mathbb{P}}
\newcommand{\E}{\mathbb{E}}

\newcommand{\rmd}{\mathrm{d}}

\newcommand{\Gcal}{\mathcal{G}}

\newcommand{\Rcal}{\mathcal{R}}

\newcommand{\Wcal}{\mathcal{W}}
\newcommand{\Xcal}{\mathcal{X}}
\newcommand{\Ycal}{\mathcal{Y}}
\newcommand{\Zcal}{\mathcal{Z}}
\newcommand{\veps}{\varepsilon}

\DeclareMathOperator\diam{diam}

\newcommand{\Sb}{\mathbf{S}}
\newcommand{\dm}{d_{\mathrm{M}}}

\newcommand{\Mut}{M}
\newcommand{\re}{\mathrm{e}}

\newtheorem{theorem}{Theorem}

\newtheorem{corollary}{Corollary}

\newtheorem{proposition}{Proposition}

\newtheorem{assumption}{\textbf{H}}

\usepackage{mathtools}
\usepackage{stackrel}

\Crefname{assumption}{\textnormal{\textbf{H}}}{\textnormal{\textbf{H}}}
\crefname{assumption}{\textnormal{\textbf{H}}}{\textnormal{\textbf{H}}}

\newcommand\numberthis{\addtocounter{equation}{1}\tag{\theequation}}

\title{Fractal Structure and Generalization Properties of Stochastic Optimization Algorithms}

\author{\name Alexander Camuto \email acamuto@turing.ac.uk \\
       \addr Department of Statistics, University of Oxford {\&} Alan Turing Institute
       \AND
       \name George Deligiannidis \email deligian@stats.ox.ac.uk\\
       \addr Department of Statistics, University of Oxford {\&} Alan Turing Institute
       \AND
       \name Murat A. Erdogdu \email  erdogdu@cs.toronto.edu \\
      \addr Department of Computer Science and Department of Statistical Sciences at University of Toronto {\&} Vector
Institute
       \AND
       \name Mert G\"{u}rb\"{u}zbalaban$^\ast$ \email mg1366@rutgers.edu \\
       \addr Department of Management Science and Information Systems, Rutgers Business School
       \AND
       \name Umut \c{S}im\c{s}ekli$^\ast$ \email umut.simsekli@inria.fr  \\
       \addr INRIA - D\'{e}partement d'Informatique de l'\'{E}cole Normale Sup\'{e}rieure - PSL Research University
       \AND
        \name Lingjiong Zhu \email zhu@math.fsu.edu\\
       \addr Department of Mathematics, Florida State University \vspace{10pt} \\
       The authors are in alphabetical order\\
       $^\ast$ Corresponding authors 
       }

\begin{document}

\maketitle

\begin{abstract}
Understanding generalization in deep learning has been one of the major challenges in statistical learning theory over the last decade.
While recent work has illustrated that the dataset and the training algorithm must be taken into account in order to obtain meaningful generalization bounds, it is still theoretically not clear which properties of the data and the algorithm determine the generalization performance.   
In this study, we approach this problem from a dynamical systems theory perspective and represent stochastic optimization algorithms as \emph{random iterated function systems} (IFS). 
Well studied in the dynamical systems literature, under mild assumptions, such IFSs can be shown to be ergodic with an invariant measure that is often supported on sets with a \emph{fractal structure}. 
As our main contribution, we prove that the generalization error of a stochastic optimization algorithm can be bounded based on the `complexity' of the fractal structure that underlies its invariant measure.
Leveraging results from dynamical systems theory, we show that the generalization error can be explicitly linked to the choice of the algorithm (e.g., stochastic gradient descent -- SGD), algorithm hyperparameters (e.g., step-size, batch-size), and the geometry of the problem (e.g., Hessian of the loss). 
We further specialize our results to specific problems (e.g., linear/logistic regression, one hidden-layered neural networks) and algorithms (e.g., SGD and preconditioned variants), and obtain analytical estimates for our bound.
For modern neural networks, we develop an efficient algorithm to compute the developed bound and support our theory with various experiments on neural networks. 
\end{abstract}

\section{Introduction}

In statistical learning,
many problems can be naturally formulated as a risk minimization problem
\begin{align}\label{eq:pop-risk}
\min\limits_{w\in\mathbb{R}^d} \Bigl\{ \mathcal{R}(w) := \E_{z\sim\pi} [\ell(\wb,z)] \Bigr\},
\end{align}
where $z \in \mathcal{Z}$ denotes a data sample coming from an unknown distribution $\pi$,
and $\ell : \mathbb{R}^d \times \mathcal{Z} \to \mathbb{R}_+$ is the composition of a loss and a function from the hypothesis class parameterized by $\wb \in \mathbb{R}^d$.
Since the distribution $\pi$ is unknown, one needs to rely on empirical risk minimization as a surrogate to \eqref{eq:pop-risk},
\begin{align}\label{eq:erm}
\min\nolimits_{w\in\mathbb{R}^d} \Bigl\{ \hat{\mathcal{R}}(\wb, \Sb_n) := ({1}/{n}) \sum\nolimits_{i=1}^n \ell(\wb,z_i) \Bigr\},
\end{align}
where $\Sb_n := \{z_1,\dots,z_n\}$ denotes a \emph{training set} of $n$ points that are independently and identically distributed (i.i.d.) and sampled 
from $\pi$, 
and model training often amounts to using an optimization algorithm to solve the above problem. 

Statistical learning theory is mainly interested in understanding the behavior of the \emph{generalization error}, i.e., $\hat{\Rcal}(w,\Sb_n) - \Rcal(w)$. While classical results suggest that models with large number of parameters should suffer from poor generalization \cite{shalev2014understanding,anthony2009neural}, modern neural networks challenge this classical wisdom: they can fit the training data perfectly, yet manage to generalize well \cite{Zhang16,neyshabur2017exploring}. Considering that the generalization error is influenced by many factors involved in the training process, the conventional algorithm- and data-agnostic uniform bounds are typically overly pessimistic in a deep learning setting.
In order to obtain meaningful, non-vacuous bounds, the underlying data distribution and the choice of the optimization algorithm need to be incorporated in the generalization bounds \cite{Zhang16,dziugaite2017computing}.

Our goal in this study is to develop novel generalization bounds that explicitly incorporate the data and the optimization dynamics, through the lens of dynamical systems theory. To motivate our approach, let us consider stochastic gradient descent (SGD), which has been one of the most popular optimization algorithms for training neural networks.
It is defined by the following recursion:
\begin{align}\label{eqn:sgd}
  &w_k = w_{k-1} - \eta \nabla \tilde{\mathcal{R}}_{k} (w_{k-1}), \quad \mbox{where} \quad \nabla\tilde{\mathcal{R}}_{k}(w) := \nabla\tilde{\mathcal{R}}_{\Omega_k}(w)  := (1/b) \sum\limits_{i\in \Omega_k} \nabla \ell(\wb,z_i). 
\nonumber
\end{align}
Here, $k$ represents the iteration counter, $\eta > 0$ is the step-size (also called the learning-rate), $\nabla \tilde{\mathcal{R}}_{k}$ is the stochastic gradient, $b$ is the batch-size, and $\Omega_k \subset \{1,2,\dots,n\}$ is a random subset drawn with or without replacement with cardinality $|\Omega_k|=b$ for all $k$.

Constant step-size SGD forms a Markov chain
with a stationary distribution $\wb_\infty\sim\mu$, which exists and is unique under mild conditions~\cite{dieuleveut2017bridging,yu2020analysis},
and intuitively we can expect that the generalization performance of the trained model to be intimately related to the behavior of the risk $\mathcal{R}(\wb)$ over this limit distribution $\mu$.
In particular, the Markov chain defined by the SGD recursion can be written by using random functions $h_{\Omega_{k}}$ at each SGD iteration $k$, i.e.,
\begin{equation} 
  w_{k} = h_{\Omega_{k}}(w_{k-1}), \ \ \text{ with }\ \
  h_{\Omega_{k}}(\wb) = \wb - \eta  \nabla \tilde{\mathcal{R}}_{k}(\wb).
\label{eq-ifs}
\end{equation}

\begin{figure}
  \centering
  \vspace{-10pt}
  \includegraphics[width=0.43\textwidth]{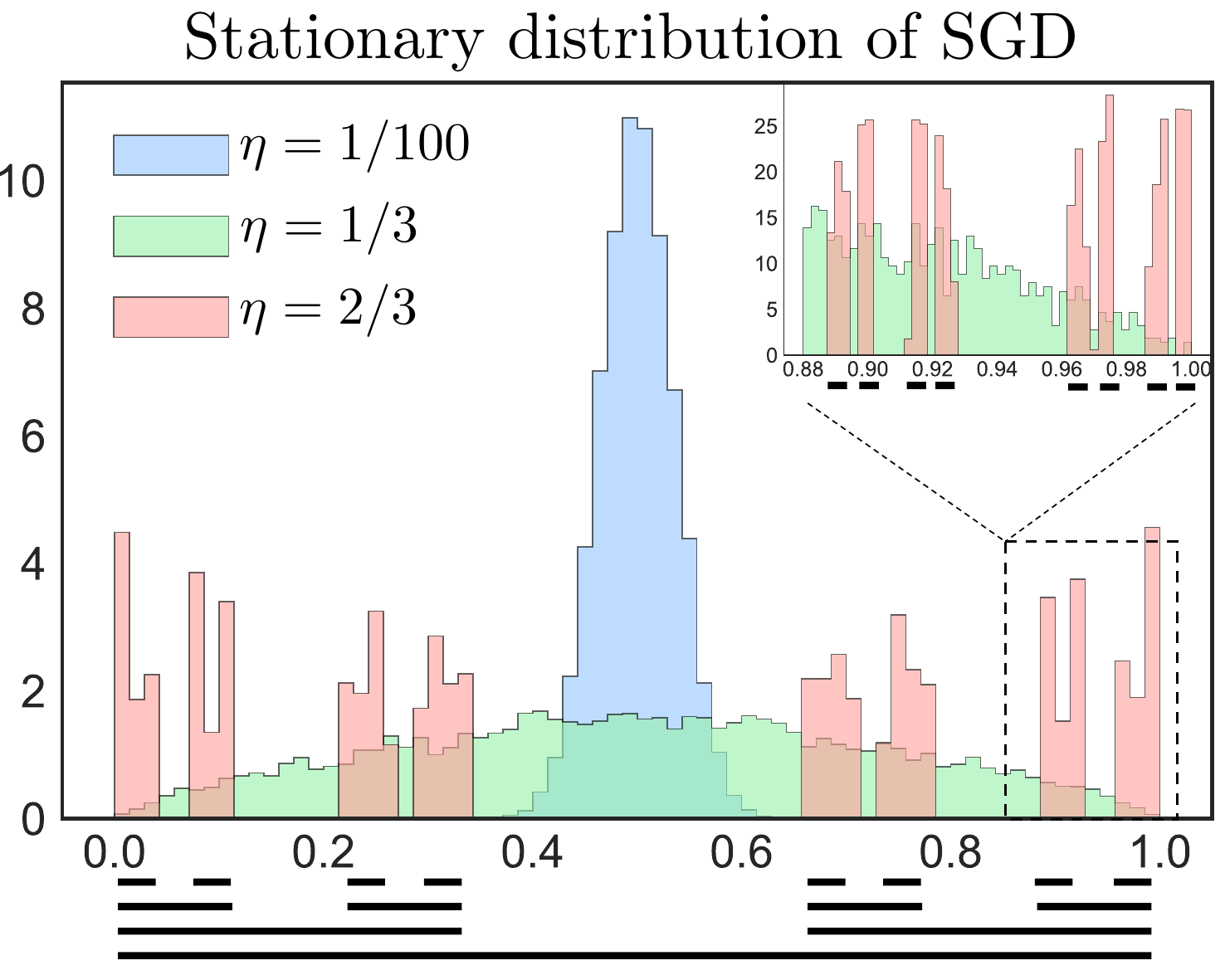}
  \caption{\small\!\! Middle-third Cantor set as the support of the stationary distribution of constant step-size SGD for $\ell(\wb,z_1) = \wb^2/2$ and
$\ell(\wb,z_2) = \wb^2/2 - \wb$.} 
  \label{fig:cantor}
  \vspace{-20pt}
\end{figure}

Here, the randomness in $h_{\Omega_{k}}$ is due to the selection of the subset $\Omega_k$.
In fact, such a formulation is not specific to SGD; it can cover many other stochastic optimization algorithms if the random function $h_{\Omega_k}$ is selected accordingly, including second-order algorithms such as 
\emph{preconditioned SGD} \cite{li2017preconditioned}.
Such random recursions \eqref{eq-ifs} and characteristics of their stationary distribution have been studied extensively 
under the names of
\emph{iterated random functions} \cite{diaconis1999iterated} and \emph{iterated function system} (IFS) \cite{falconer2004fractal}.
In this paper, from a high level, we relate the `complexity of the stationary distribution' of a particular IFS to the generalization performance of the trained model.

We illustrate our context in two toy examples.
In the first one, we consider a 1-dimensional quadratic problem with $n=2$ and $\ell(\wb,z_1) = \wb^2/2$ and
$\ell(\wb,z_2) = \wb^2/2 - \wb$. We run SGD with constant step-size $\eta$ to minimize the resulting empirical risk. We simply choose $\Omega_k \subset\{1,2\}$ uniformly random with batch-size $b=1$, and we plot the histograms of stationary distributions for different step-size choices $\eta \in \{1/100,1/3,2/3 \}$ in Figure~\ref{fig:cantor}.
We observe that the \emph{support} of the stationary distribution of SGD depends on the step-size:  
As the step-size increases the support becomes less dense and a \emph{fractal structure} in the stationary distribution can be clearly observed. 
This behavior is not surprising, at least for this toy example. It is well-known that the set of points that is invariant under the resulting IFS (termed as the attractor of the IFS) for the specific choice of $\eta =2/3$ is the famous `middle-third Cantor set'~\cite{feng2009structures}, which coincides with the support of the stationary distribution of the SGD.

\begin{figure}[t]
\centering

\subfigure[$\eta=0.3$]{\includegraphics[width=0.2\textwidth]{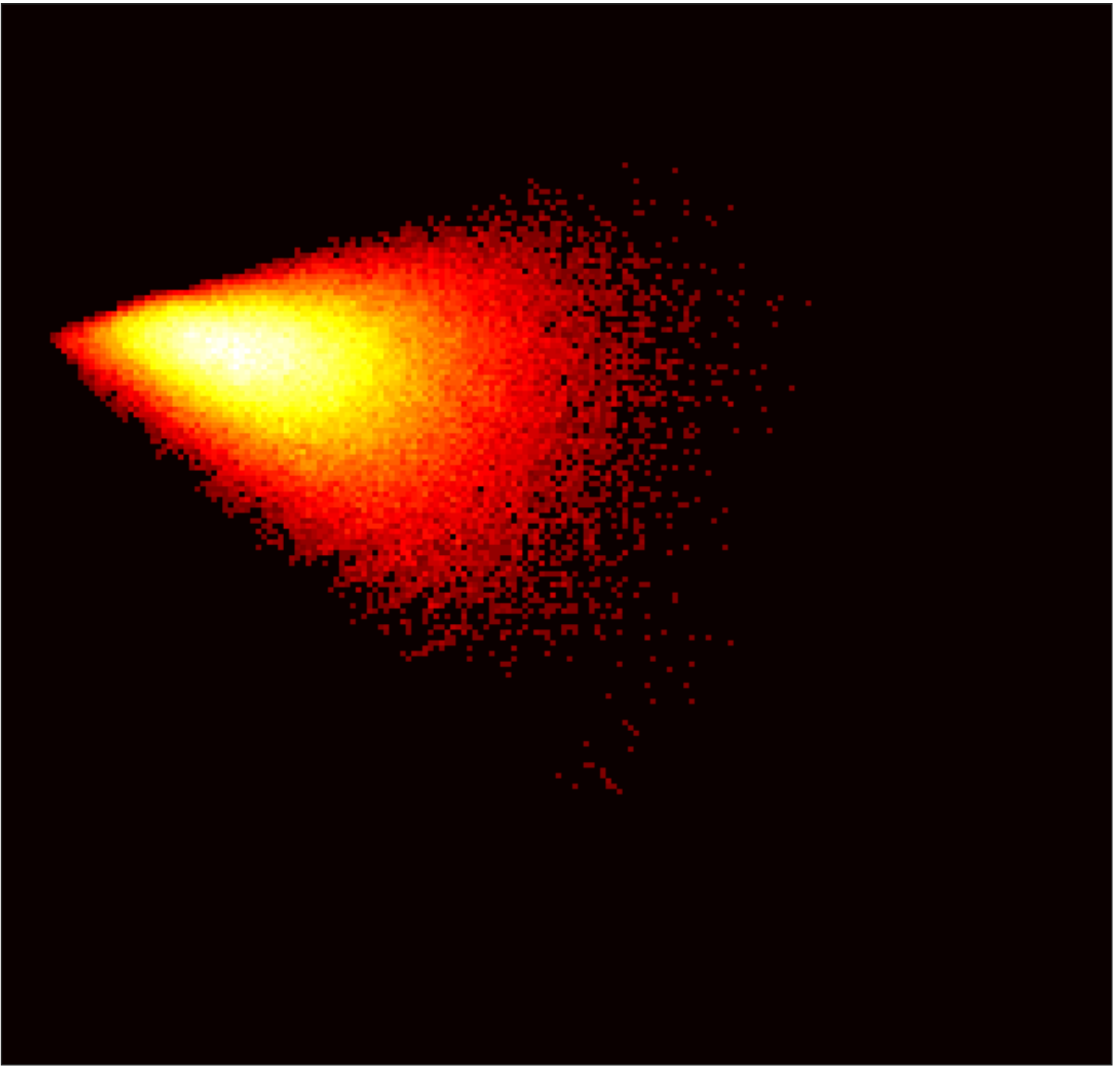}}
\subfigure[$\eta=0.5$]{\includegraphics[width=0.2\textwidth]{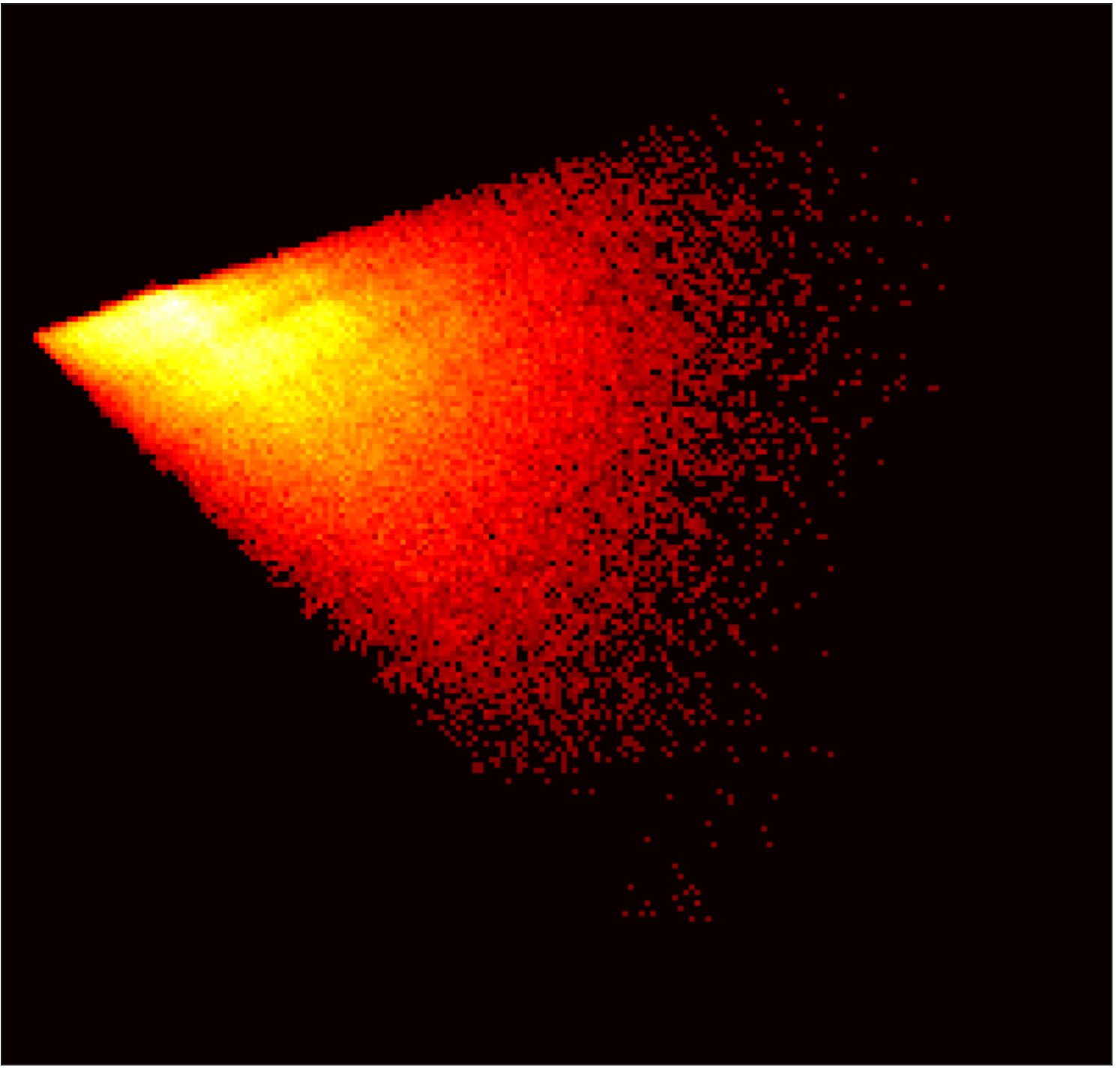}}
\subfigure[$\eta=0.7$]{\includegraphics[width=0.2\textwidth]{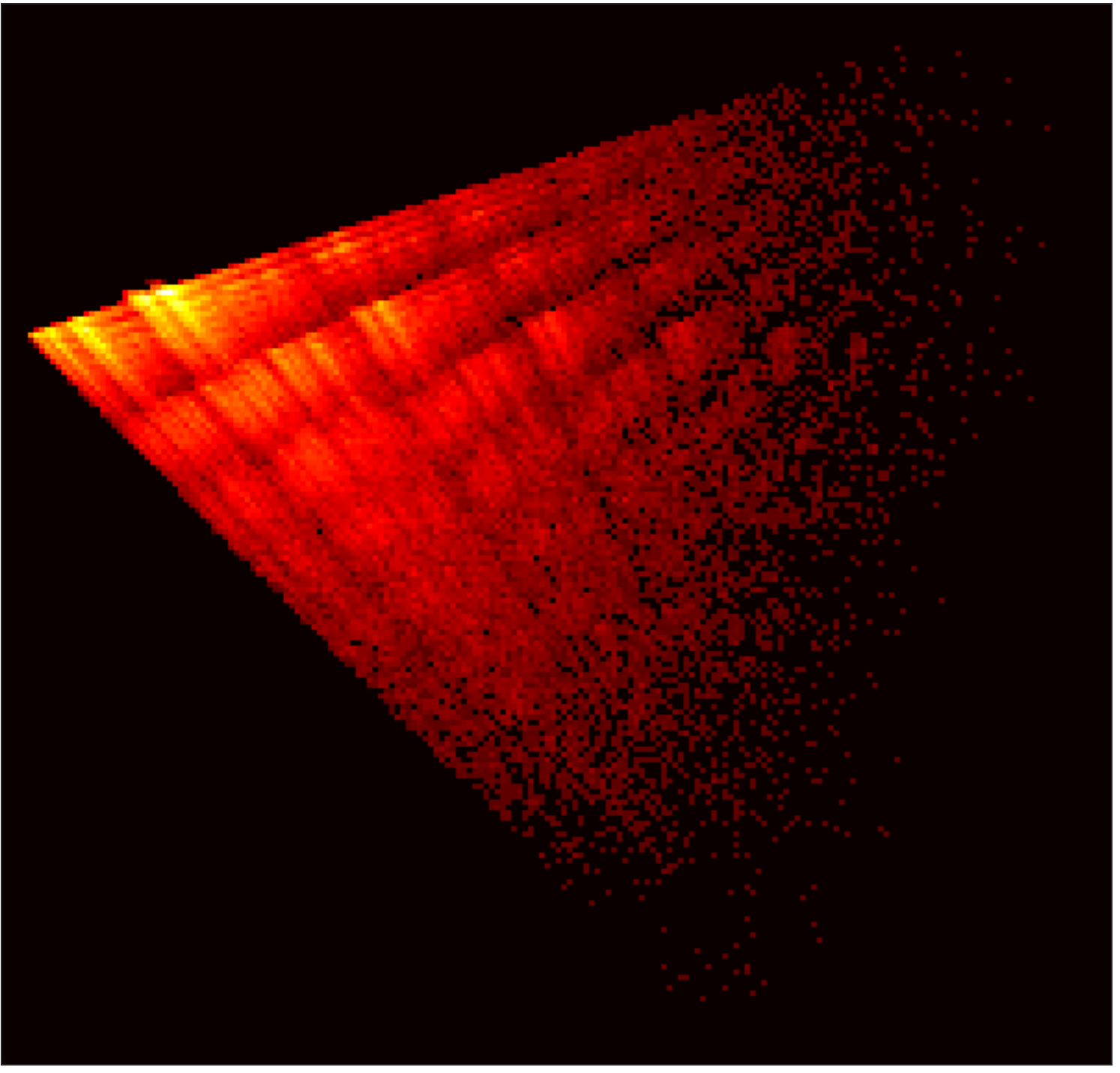}}
\subfigure[$\eta=0.9$]{\includegraphics[width=0.2\textwidth]{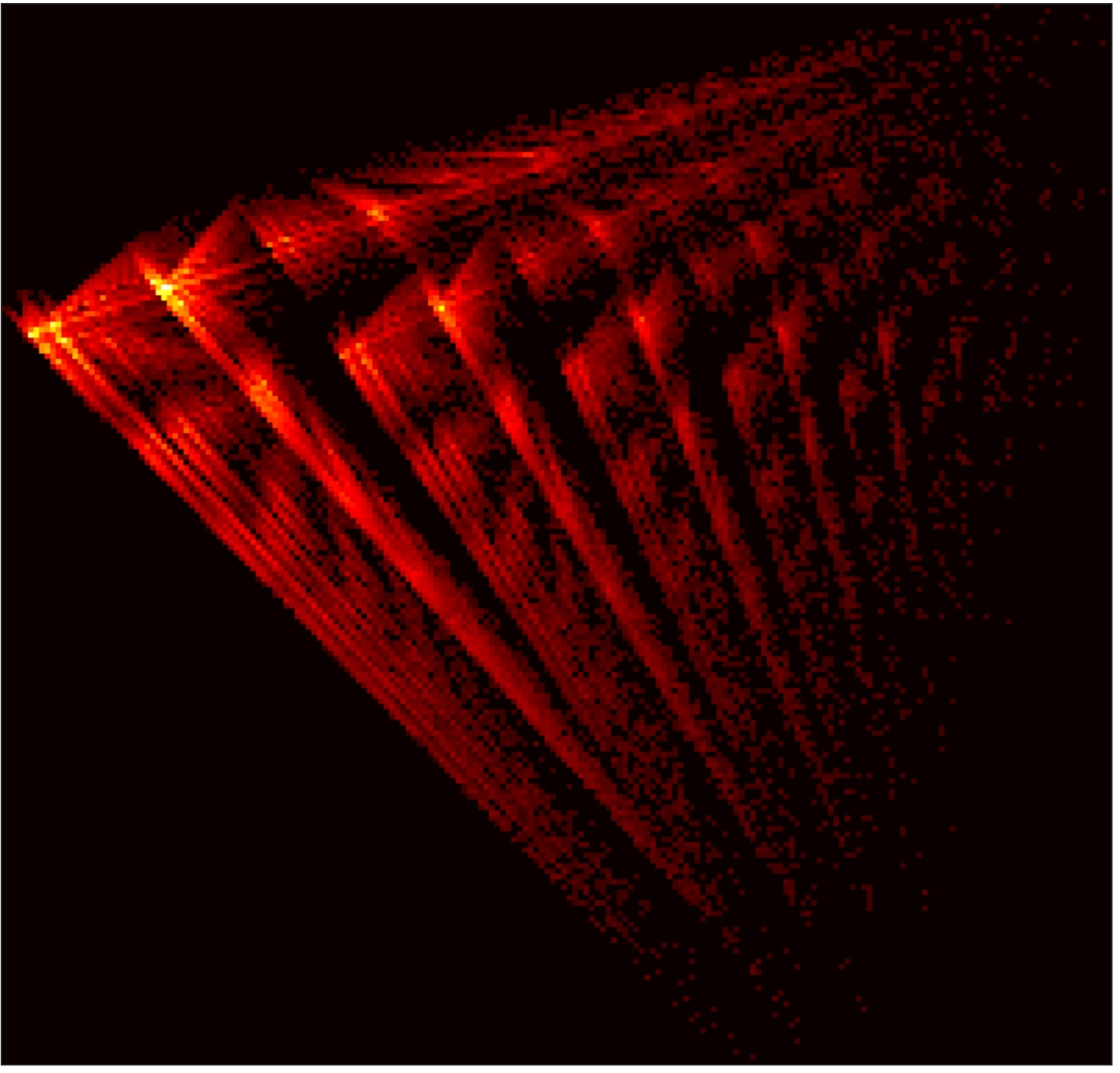}}

\caption{The stationary distribution of constant step-size SGD for linear regression, where we have $n=5$ data points and $w \in \rset^2$. }
\label{fig:example}
\vspace{-3mm}
\end{figure}

As another example, we run SGD with constant step-size in order to train an ordinary linear regression model for a dataset of $n=5$
samples and $d=2$ dimensions, i.e., $a_i^\top w \approx y_i$, where for $i=1,\dots,5$, $y_i$ and each coordinate of $a_i$ are drawn uniformly at random from the interval $[-1,1]$. Figure~\ref{fig:example} shows the heatmap of the resulting stationary distributions for different step-size choices $\eta$ ranging from $0.1$ to $0.9$ (bright colors represent higher density).
We observe that for small step-size choices, the stationary distribution is dense, whereas a fractal structure can be clearly observed as the step-size gets larger.

Fractals are complex patterns and the level of this complexity is typically measured by the \emph{Hausdorff dimension} 
of the fractal, which is a notion of dimension that can take fractional values\footnote{The Hausdorff dimension of the middle-third Cantor set in Figure~\ref{fig:cantor} is $\log_3(2) \approx 0.63$
whereas the ambient dimension is $1$~\cite[Example 2.3]{falconer2004fractal}.},
and can be much smaller than the ambient dimension $d$. 
Recently, assuming that {SGD trajectories} 
can be well-approximated by a certain type of stochastic differential equations (SDE), 
it is shown that the generalization error can be controlled by the Hausdorff dimension
of the trajectories of the SDE, instead of their ambient dimension $d$~\cite{simsekli2020hausdorff}.
That is, the ambient dimension that appears in classical learning theory bounds is replaced with the Hausdorff dimension.
The fractal geometric approach presented in \cite{simsekli2020hausdorff} can capture the `low dimensional structure' of fractal sets and provides an alternative perspective to the compression-based approaches that aim to understand why overparametrized networks do not overfit \cite{arora2018stronger,Suzuki2020Compression,suzuki2018spectral,hsu2021generalization}.

However, SDE approximations for SGD 
often serve as mere heuristics, and guaranteeing a good approximation typically requires unrealistically small step-sizes \cite{li_jmlr2019}. For more realistic step-sizes, theoretical concerns have been raised about the validity of conventional SDE approximations for SGD \cite{li2021validity,ht-phenomenon,yaida2018fluctuationdissipation}. Another drawback of the SDE approximation is that the bounds in \cite{simsekli2020hausdorff} are implicit, in the sense that they cannot be related to algorithm hyperparameters, problem geometry, or data.
We address these issues and present a direct, \emph{discrete-time} analysis by exploiting the connections between IFSs and stochastic optimization algorithms. 

Our contributions are summarized as follows:
\begin{itemize}[noitemsep,topsep=0pt,leftmargin=.11in] 
    \item We extend \cite{simsekli2020hausdorff} and show that the generalization error can be linked to the Hausdorff dimension of \emph{invariant measures} (rather than the Hausdorff dimension of \emph{sets} as in \cite{simsekli2020hausdorff}). More precisely, under
    appropriate conditions,
    we establish a generalization bound for the stationary distribution of IFS $w_\infty\sim \mu$.
    That is, with probability at least $1-2\zeta$,
\begin{align}
    |\hat{\Rcal}(w_\infty,\Sb_n) - \Rcal(w_\infty)| \lesssim \sqrt{\frac{ \dimhu \mu \, \log^2 (n)}{n} + \frac{\log(1/\zeta)}{n}},
\end{align}
 for $n$ large enough, where $\dimhu \mu$ is the (upper) Hausdorff dimension of the measure $\mu$. \vspace{2pt}

 \item By leveraging results from IFS theory, we further link $\dimhu \mu$ to (i) the form of the recursion (e.g., $h_{\Omega_k}$ in \eqref{eq-ifs}), (ii) algorithm hyperparameters (e.g., $\eta$, $b$), and (iii) problem geometry (e.g., Hessian of $\tilde{\Rcal}_k$), through a single term, which encapsulates all these components and their interaction. \vspace{2pt}
 \item We establish bounds on $\dimhu \mu$ for SGD and preconditioned SGD algorithms, when used to minimize various empirical risk minimization problems
 such as least squares, logistic regression, support vector machines.
 In all cases, we explicitly link the generalization performance of the model to the hyperparameters of the underlying training algorithm. 

 \vspace{2pt}
 \item Finally, we numerically compute key quantities that appear in our generalization bounds, and show empirically that they have a statistically significant correlation with the generalization error.
\end{itemize}

\textbf{Notation and preliminaries. } 
$B_d(x,r) \subset \rset^d$ denotes the closed ball centered around $x \in \rset^d$ with radius $r$.

A function $f : \mathbb{R}^{d_1} \to \mathbb{R}^{d_2}$ is said to be (Fr\'echet) differentiable at $x\in \mathbb{R}^{d_1}$ if there exists a $d_1\times d_2$ matrix ${\displaystyle J_{f}(x):\mathbb{R}^{d_1}\to \mathbb{R}^{d_2}}$ such that
$\lim _{\|h\|\to 0}\|f(x+h)-f(x)-J_{f}(x)h\|/\|h\|=0$.
The matrix $J_{f}(x)$ is called the differential of $f$,
also known as the Jacobian matrix at $x$,
and determinant of $J_{f}(x)$ is called the Jacobian determinant \cite{HS1974}.
For real-valued functions $f,g$, we define $f(n)=\omega(g(n))$ if $\lim_{n\rightarrow\infty}|f(n)|/g(n)=\infty$. For a set $A$, $|A|$ denotes its cardinality.

\section{Technical Background on Fractal Geometry}

\label{sec:background}

Fractal sets emerge virtually in all branches of science, and fractal-based techniques have been used in machine learning \cite{sugiyama2013learning,malach2019deeper,dym2019expression,simsekli2020hausdorff, agarwal2021acceleration}.
The inherent `complexity' of a fractal set often plays an important role and it is typically measured by its \emph{fractal dimension}, where several notions of dimension have been proposed \cite{falconer2004fractal}. In this section, we  briefly mention two important notions of fractal dimension, which will 
be used in our 
theoretical development. 

\textbf{Minkowski dimension of a set. } 
The Minkowski dimension (also known as the box-counting dimension \cite{falconer2004fractal}) is defined as follows.
Let $F \subset \mathbb{R}^d$ be a set and for $\delta>0$, let $N_\delta(F)$ denote a collection of sets that contains the smallest number of closed balls of diameter at most $\delta$ which cover $F$.
Then the upper-Minkowski dimension of $F$ is defined as follows:
\begin{equation}\label{eqn:dimmink}
\dimmu F := \limsup\limits_{\delta \to 0} \Bigl[ {\log |N_\delta(F)|}\>/\>{\log (1/\delta)} \Bigr]. 
\end{equation}
To visualize the upper-Minkowski dimension of a set $F$, 
consider the set $F$ lying on an evenly spaced grid
and count how many boxes are required to cover the set.
The upper-Minkowski dimension measures how this number changes as the grid is made finer using a box-counting algorithm.

\textbf{Hausdorff dimension of a set. }
An alternative to the purely geometric Minkowski dimension, the Hausdorff dimension \cite{Hausdorff1918} is a measure theoretical notion of fractal dimension. 
It is based on the \emph{Hausdorff measure}, which generalizes the traditional notions
of area and volume to non-integer dimensions \cite{rogers1998}. More precisely, for $s\geq 0$, let $F\subset\mathbb{R}^{d}$ and $\delta>0$, and denote $\mathcal{H}_{\delta}^{s}(F):=\inf\sum_{i=1}^{\infty}\diam(A_{i})^{s}$,
where the infimum is taken over all the $\delta$-coverings
$\{A_{i}\}_{i}$ of $F$, that is, $F\subset\cup_{i}A_{i}$
with $\diam(A_{i})<\delta$ for every $i$.
The $s$-dimensional {Hausdorff measure} of $F$
is defined as the monotone limit $\mathcal{H}^{s}(F):=\lim_{\delta\rightarrow 0}\mathcal{H}_{\delta}^{s}(F)$. 
When $s\in\mathbb{N}$, $\mathcal{H}^{s}$ is the $s$-dimensional
Lebesgue measure up to a constant factor; hence the generalization of `volume' to fractional orders.

Based on the Hausdorff measure, the \emph{Hausdorff dimension} of a set $F\subset\mathbb{R}^{d}$ is then defined as follows:
\begin{equation*}
\dimh F:=\sup\left\{s>0:\mathcal{H}^{s}(F)>0\right\}
=\inf\left\{s>0:\mathcal{H}^{s}(F)<\infty\right\}.
\end{equation*}
In other words, the Hausdorff dimension of $F$ is the `moment' $s$ when $\mathcal{H}^{s}(F)$ drops from $\infty$ to $0$, that is, $\mathcal{H}^{r}(F)=0$
for all $r>\dimh F$
and $\mathcal{H}^{r}(F)=\infty$
for all $r<\dimh F$.

We always have $0\leq \dimh F \leq d$, and when $F$ is bounded, we always have $0\leq\dimh F \leq \dimmu F\leq d$ \cite{falconer2004fractal}. Furthermore, the Hausdorff dimension of $\mathbb{R}^{d}$
equals $d$, 
and the Hausdorff dimension of smooth Riemannian manifolds
correspond to their intrinsic dimension, e.g. $\dimh \mathbb{S}^{d-1}=d-1$, where $\mathbb{S}^{d-1}$ is the unit sphere in $\mathbb{R}^d$.

\textbf{Hausdorff dimension of a probability measure. }
IFSs generate invariant measures as the number of iterates goes to infinity,
and random fractals arise from such invariant measures.
There has been a growing literature
that studies the structure of such random fractals \cite{Szarek,NSB2002,MS2002,rams2006dimension,FST2006,JR2008}, where
the notion of fractal dimension can be extended to measures, and our theory will rely on the Hausdorff dimension of invariant measures associated with stochastic optimization algorithms.
In particular, we will mainly use the \emph{upper Hausdorff dimension} $\dimhu \mu$ of a Borel probability measure $\mu$ on $\mathbb{R}^d$, which is defined as follows:
$\dimhu \mu := \inf \left\{\dimh A : \mu(A) = 1 \right\}$.
In other words, $\dimhu \mu$ is the smallest Hausdorff dimension of all measurable sets with full measure.

\section{Generalization Bounds for Stochastic Optimization Algorithms as IFSs} 

\label{sec:generalization}

In this section, we will present our main theoretical results which relate the generalization error to the upper-Hausdorff dimension of the invariant measure associated with a stochastic optimization algorithm. 
We consider a standard supervised learning setting, where $\mathcal{Z}=\mathcal{X}\times \mathcal{Y}$, where $\Xcal$ is the space of features and $\Ycal$ is the space of labels, and $\pi$ is the unknown data distribution on $\Zcal$.  

For mathematical convenience, in order to construct the training set with $n$ elements, we first consider an \emph{infinite sequence} of i.i.d.\ data samples from the data distribution $\pi$, then we take the first $n$ elements from this infinite sequence. More precisely, we consider the (countable) product measure $\pi^{\infty} = (\pi \otimes \pi \otimes \dots)$ defined on the cylindrical sigma-algebra. Then, we consider the infinite i.i.d.\ data sequence as $\mathbf{S} \sim \pi^\infty$, i.e., $\mathbf{S} = (z_j)_{j\geq 1}$ with $z_j \stackrel{\mathclap{\mbox{\tiny{i.i.d.}}}}{\sim} \pi$ for all $j = 1,2, \dots$. 
Finally, we define the training set as $\mathbf{S}_n := (z_1, \dots, z_n)$, i.e., we take the first $n$ elements of $\mathbf{S}$. 
To avoid technical complications, throughout the paper we will assume that all the encountered functions and sets are measurable. All the proofs are given in the supplement.

Given a dataset $\Sb_n$, we represent the \emph{training algorithm} as an IFS, which is based on the following recursion:
$w_k = h_{\Omega_k}(w_{k-1}; \Sb_n)$, where the mini-batch $\Omega_k$ is i.i.d.\ sampled according to some distribution (e.g.,  sampling without-replacement uniformly among all possible mini-batches). 
This compact representation enables us to cover a broad range of optimization algorithms  with a unified notation, including SGD (see \eqref{eq-ifs}), as well as preconditioned SGD, and stochastic Newton methods. For example, if we take $ h_{\Omega_k}(w; \Sb_n) = w - \eta  {H}_k(w)^{-1}\nabla \tilde{\Rcal}_k(w)$ where ${H}_k(w)$ is an estimate of the Hessian of $\tilde{\Rcal}_k$, we cover stochastic Newton methods \cite{erdogdu2015convergence}. Similar constructions can be made for other popular algorithms, such as SGD-momentum \cite{qian1999momentum}, RMSProp \cite{hinton2012overview}, or Adam \cite{kingma2014adam}. 

Notice that there are only finitely many values that $\Omega_k$ can take. For example, in the case of without-replacement mini-batch sampling with batch-size $b$, there are in total $m_b=\binom{n}{b}$ many subsets of $\{1,2,\dots, n\}$ with cardinality $b$. Alternatively, a more realistic setup would be to divide the dataset into $m_b = n/b$ batches with each batch having $b$ elements, and at each iteration $k$, we can randomly choose one of the batches. In both examples we can enumerate as $S_1, S_2, \dots, S_{m_b}$. If the probability of sampling the mini-batch $\Omega_k = S_i$ is $p_i$ for every $i$, then, with a slight abuse of notation, we can rewrite the IFS recursion as:
\begin{equation} 
w_k = h_{U_k}(w_{k-1}; \Sb_n),
\label{eq-ifs-2}
\end{equation}
where $U_k$ is a random variable taking values in $\{1,2,\dots, m_{b}\}$ and $p_i:=\mathbb{P}(U_k = i)$. If the mini-batch sampling is uniform (i.e., the default option in practice), we have $p_i= 1/m_{b}$; however, we are not restricted to this option, the sampling scheme is allowed to be more general. We finally call the triple $(\mathbb{R}^d, \{h_i(\cdot; \Sb_n)\}_{i=1}^{m_b},  \{p_i\}_{i=1}^{m_b})$ an iterated function system (IFS). 

Given a dataset $\Sb_n$, we are interested in the limiting behavior of the training algorithm \eqref{eq-ifs-2}. We characterize this behavior by considering the invariant measure $\mu_{W|\Sb_n}$ of the IFS (also called stationary distribution), that is a Borel probability measure on $\rset^d$, such that $w_{\infty} \sim \mu_{W|\Sb_n}$. 
To be able to work in this context, we first need to ensure that the recursion \eqref{eq-ifs-2} admits an invariant measure, i.e., $\mu_{W|\Sb_n}$ exists. Accordingly, we require the following mild conditions on the IFS \eqref{eq-ifs-2}. 
Let $U$ be a random variable with the same distribution as $U_k$. If the recursion \eqref{eq-ifs-2} is \emph{Lipschitz on average}, i.e.,  
\begin{align} 
\mathbb{E}[L_U \>|\> \Sb_n ] < \infty, \quad \mbox{with} \quad L_U := \sup\limits_{x,y\in\mathbb{R}^d} \frac{\|h_{U}(x; \Sb_n) - h_{U}(y; \Sb_n) \|}{\|x-y\|},
\label{assump-lip}
\end{align}
and is \emph{contractive on average}, i.e., if 
\begin{equation} 
\mathbb{E}\left[\log(L_U) \>|\> \Sb_n \right] = \sum\limits_{i=1}^{m_{b}} 
p_i \log(L_i) <0,  \qquad \text{with \phantom{aa} $p_i>0$, for any $i= 1,\dots,m_b$},
\label{assump-mean-contractive}
\end{equation} 
then it can be shown
that the process is ergodic and admits a unique invariant measure where the limit
$$ \rho := \lim\limits_{k\to\infty} (1/k)\log \left\|h_{U_k} h_{U_{k-1}} \cdots h_{U_1}\right\|$$ 
exists almost surely and is a constant \cite{elton1990multiplicative}, where $\rho$ is called the \emph{Lyapunov exponent}. Furthermore, under further technical assumptions, it can be shown that \eqref{eq-ifs-2} is geometrically ergodic \cite{diaconis1999iterated}.

Our goal will be to relate the generalization error to $\dimhu \mu_{W|\Sb_n}$. To achieve this goal, at first sight, it might seem tempting to extract a full-measure set $A$ by using the definition of $\mu_{W|\Sb_n}$, such that $\mu_{W|\Sb_n}(A) = 1$ and $\dimh A \approx \dimhu \mu_{W|\Sb_n}$, and then directly invoke the results from \cite{simsekli2020hausdorff}, which would link the generalization error to $\dimh A$, hence, also to $\dimhu \mu_{W|\Sb_n}$. However, since \cite{simsekli2020hausdorff} does not consider an IFS framework, the conditions they require (e.g., boundedness of $A$, $\dimm A = \dimh A$) are not suited to IFSs, and hence prevent us from directly using their results. 

As a remedy, we make a detour and show that we can find \emph{almost full-measure} sets $A$, such that $\mu_{W|\Sb_n}(A) \approx 1$ and $\dimmu A \approx \dimhu \mu_{W|\Sb_n}$ (notice that in this case we directly use the Minkowski dimension of $A$, as opposed to its Hausdorff dimension).
To achieve this goal, we require the following geometric regularity condition on the invariant measure. 
\begin{assumption}
\label{asmp:localreg}
For $\pi^\infty$-almost all $\Sb$ and all $n \in \mathbb{N}_+$, the recursion \eqref{eq-ifs-2} satisfies \eqref{assump-lip} and \eqref{assump-mean-contractive} and the limit \[\lim_{r \to 0}\Bigl[ {\log \mu_{W|\Sb_n}(B_d(w,r))}\>/\>{\log r}\Bigr]\]
exists for $\mu_{W|\Sb_n}$-almost every $w$. 
\end{assumption} 

This is a common condition \cite{pesin2008dimension} and is satisfied for a large class of measures. For instance, `sufficiently regular' measures with the property that $ C_1 r^s \leq \mu(B(x,r)) \leq C_2 r^s$ for some constant $s>0$ and positive constants $C_1$, $C_2$ will satisfy this assumption. Such measures are called Ahlfors-regular (cf.\ \cite[Assumption H4]{simsekli2020hausdorff} for a related condition), and
it is known that IFSs that satisfy certain `open set conditions' lead to Ahlfors regular invariant measures (see \cite[Section 8.3]{mackay2010conformal}). Yet, our assumption is more general and does not immediately require Ahlfors-regularity. 
Under \Cref{asmp:localreg}, we now formalize our key observation, which serves as the basis for our bounds.
\begin{proposition}
\label{lem:existence}

Assume that \Cref{asmp:localreg} holds. Then for every $\veps >0$, $n\in \mathbb{N}_+$, and $\pi^\infty$-almost every $\Sb$, there exists $\delta := \delta(\veps,\Sb_n) \in (0,1]$ and a bounded measurable set $A_{\Sb_n,\delta} \subset \rset^d$, such that
\begin{align}
\label{eqn:regsets}
    \mu_{W|\Sb_n}(A_{\Sb_n,\delta}) \geq 1-\delta, \quad \text{and} \quad \dimmu A_{\Sb_n,\delta} \leq \dimhu \mu_{W|\Sb_n} + \veps,
\end{align}
and $\delta(\veps,\Sb_n) \to 0$ as $\veps \to 0$.
\end{proposition}

Thanks to this result, we can now leverage the proof technique presented in \cite[Theorem 2]{simsekli2020hausdorff}, and link the generalization error to $\dimhu \mu_{W|\Sb_n}$ through $\dimmu A_{\Sb_n,\delta}$.
We shall emphasize that, mainly due to the sets $A_{\Sb_n,\delta}$ not being of full-measure, our framework introduces additional non-trivial technical difficulties that we need to tackle in our proof.

We now introduce our second assumption, which roughly corresponds to a `topological stability' condition, and is adapted from \ \cite[Assumption H5]{simsekli2020hausdorff}. Formally, consider the (countably infinite) collection of closed balls of radius $\beta$, whose centers are on the fixed grid 
$N_\beta := \left\{\left( \frac{(2j_1+1) \beta}{2\sqrt{d}},\dots,\frac{(2j_d+1) \beta}{2\sqrt{d}} \right): j_i \in \mathbb{Z}, i=1,\dots,d\right\}$,
and for a set $A \subset \rset^d$, define $N_\beta(A):= \{x\in N_{\beta}: B_d(x,\beta) \cap A \neq \emptyset\}$, which is the collection of the centers of the balls that intersect $A$.

\begin{assumption}\label{asmp:decoupling2}

  Let $\Zcal^\infty := (\Zcal \times \Zcal \times \cdots)$ denote the countable product endowed with the product topology and let $\mathfrak{B}$ be the Borel $\sigma$-algebra generated by $\Zcal^\infty$.
For a Borel set $A \subset \rset^d$, let $\mathfrak{F}, \mathfrak{G}$ be the sub-$\sigma$-algebras of $\mathfrak{B}$ generated by the collections of random variables given by
$\{ \hat{\Rcal}(w,\Sb_n): w \in \rset^d, n \geq 1\}$ and
$\Big\{ \mathds{1}\left\{w\in N_{\beta}(A)\right\}: \beta\in \mathbb{Q}_{>0}, w\in N_\beta, n \geq 1 \Big\}$ respectively.
There exists a constant $\Mut \geq 1$ such that for any $F\in \mathfrak{F}$, $G\in \mathfrak{G}$ we have $\pr\left( F \cap G\right) \leq \Mut \pr\left( F \right) \pr(G)$ for all Borel sets $A$.
\end{assumption}
\Cref{asmp:decoupling2} simply ensures that the dependence between the training error and the topological properties of the support of $\mu_{W|\Sb_n}$ can be controlled via $M$. Hence, it can be seen as a form of \emph{algorithmic stability} \cite{bousquet2002stability}, where $M$ measures the level of stability of the topology of $\mu_{W|\Sb_n}$: a small $M$ indicates that the geometrical structure of $\mu_{W|\Sb_n}$ does not heavily depend on the particular value of $\Sb_n$. 
The constant $M$ is also related to the mutual information \cite{xu2017information,asadi2018chaining}, 
but may be better behaved than the mutual information as it relies on very specific functions of the random variables.

We require one final assumption, which states that the loss $\ell$ is sub-exponential. 
\begin{assumption}\label{asmp:subexp}
$\ell$ is $L$-Lipschitz continuous in $w$, and when $z\sim \pi$, for all $w$, $\ell(w,z)$ is $(\nu, \kappa)$-sub-exponential, that is, for all $|\lambda|<1/\kappa$,
we have $\log \mathbb{E}_{z\sim \pi}\left[\exp\left(\lambda\left(\ell(w,z)-\mathcal{R}(w) \right)\right)\right]
\leq \nu^2\lambda^2/\kappa$.
\end{assumption}

Armed with these assumptions, we can now present our main result.

\begin{theorem}
\label{thm:generalization}
Assume that \Cref{asmp:localreg,asmp:decoupling2,asmp:subexp} hold 
and $\dimhu \mu_{W|\Sb_n} = \omega(\log\log(n)/\log (n)) $, $\pi^\infty$-almost-surely. Then, the following bound holds for sufficiently large $n$:
\begin{align}
    |\hat{\Rcal}(W,\Sb_n) - \Rcal(W)| \leq 8\nu \sqrt{\frac{ \dimhu \mu_{W|\Sb_n}  \log^2 (nL^2)}{n} + \frac{\log(13M/\zeta)}{n}},
\end{align}
with probability at least $1-2\zeta$ over the joint distribution of $\Sb \sim \pi^{\infty}$, $W \sim \mu_{W|\Sb_n}$.
\end{theorem}

This theorem shows that the Hausdorff dimension of the invariant measure acts as a `capacity metric' and the generalization error is therefore directly linked to this metric, i.e., the complexity of the underlying fractal structure has close links to the generalization performance. On the other hand, the condition $\dimhu \mu_{W|\Sb_n} = \omega(\log\log(n)/\log (n)) $ is very mild and makes sure that the dimension of the IFS does not decrease very rapidly with increasing number of data 
points $n$. 
Theorem~\ref{thm:generalization} enables us to access the rich theory of IFSs, where bounds on the Hausdorff dimension are readily available, and connect them to statistical learning theory. The following result is a direct corollary to Theorem~\ref{thm:generalization} and \cite[Theorem 2.1]{rams2006dimension} (see Theorem~\ref{thm-rams} in the Appendix).
\begin{corollary}
\label{coro-main-thm}
Assume that the conditions of Theorem~\ref{thm:generalization} hold. Furthermore, consider the recursion \eqref{eq-ifs-2} and assume that $h_i$ are continuously differentiable with derivatives $h'_i$ that are $\alpha$-H\"older continuous for some $\alpha>0$. 
Then, there exists a constant $M>1$ such that 
for sufficiently large $n$:
\begin{align}
    |\hat{\Rcal}(W,\Sb_n) - \Rcal(W)| \leq 8 \nu \sqrt{ \frac{\mathrm{Ent}\log^2 (nL^2)}{\left[\sum\limits_{i=1}^{m_{b}} p_i \int_{\mathbb{R}^d} \log(\|J_{h_{i}}(w)\|) \rmd\mu_{W|\Sb_n}(w)\right]n }   + \frac{\log(13M/\zeta)}{n}},
\end{align}
with probability $1-2\zeta$ over $\Sb \sim \pi^{\infty}$, $W \sim \mu_{W|\Sb_n}$, where $\mathrm{Ent} := \sum_{i=1}^{m_{b}} p_i \log(p_i)$ denotes the negative entropy of the mini-batch sampling scheme, $\|\cdot\|$ denotes the operator norm, and $J_{h_i}$ is the Jacobian of $h_i$ defined in the notation section.

\end{corollary}
By this result, we discover an interesting quantity, $\sum\nolimits_{i=1}^{m_{b}} p_i \int_{\mathbb{R}^d} \log(\|J_{h_{i}}(w)\|) \rmd\mu_{W|\Sb_n}(w)$, which \emph{simultaneously} captures the effects of the data and the algorithm.
To see it more clearly, let us consider the SGD recursion \eqref{eq-ifs}, where $\|J_{h_i}(w) \| = \|I - \eta \nabla^2 \tilde{\Rcal}_{S_i}(w)\|$ and $\{S_i\}_{i=1}^{m_b}$ denotes the enumeration of the mini-batches. Then, the overall quantity becomes
\begin{align}
\label{eqn:newterm}
\mathbb{E}_{U, W} \bigl[ \log\|I - \eta \nabla^2 \tilde{\Rcal}_{S_U}(W)\|  \bigr],    
\end{align}
where the expectation is taken over the mini-batch index $U\in \{1,\dots,m_b\}$ with $\mathbb{P}(U = i) = p_i$, and $W \sim \mu_{W|\Sb_n}$. We clearly observe that this term depends on (i) the algorithm choice through the form of $h_i$, (ii) step-size $\eta$, (iii) batch-size through $m_b$, (iv) problem geometry through $\nabla^2 \tilde{\Rcal}$, and (v) data distribution through $\mu_{W|\Sb_n}$. We believe that such a compact representation of all these constituents and their interaction is novel and will open up interesting future directions.

\section{Analytical Estimates for the Hausdorff Dimension}
\label{sec:examples}

The generalization bound presented in Theorem~\ref{thm:generalization} applies to a number of stochastic optimization algorithms that can be represented with an IFS and to a large class of losses that can be non-convex or convex. It is controlled by the Hausdorff dimension of the invariant measure $\mu_{W|\Sb_n}$ which needs to be estimated. In the numerical experiments section, we will discuss how this quantity can be estimated from the dataset $\Sb_n$ and the iterates of the underlying algorithm.

Corollary~\ref{coro-main-thm} shows that for smooth losses, the Hausdorff dimension can be controlled with the expectation of the norm of the logarithm of the Jacobian $\log(\|J_{h_i}(w)\|)$ with respect to the invariant measure $\mu_{W|\Sb_n}$. In general, an explicit characterization of the invariant measure is not known. Nevertheless, under additional appropriate assumptions that can hold in practice, such as boundedness of the data of the loss, we next discuss that it is possible to get uniform lower and upper bounds on the quantity $\|J_{h_i}(w)\|$ which leads to analytical upper bounds on $\dimhu \mu_{W|\Sb_n}$. 

As illustrative examples; in the following, we will consider the setting where we divide $\Sb_n$ into $m_b = n/b$ batches with each batch having $b$ elements, and then we discuss how analytical estimates on the (upper) Hausdorff dimension $\dimhu \mu_{W|\Sb_n}$ can be obtained for some particular problems 
including least squares, regularized logistic regression, and one hidden-layer networks.
In the Appendix, we also discuss how similar bounds can be obtained for support vector machines and other algorithms such as preconditioned SGD and stochastic Newton methods.

\textbf{Least squares. }
We consider the least squares problem, with data points $z_i = (a_i, y_i)$ and loss
\begin{align}
\label{pbm-lse}
    \ell(w,z_i) :=  \left(a_i^Tw - y_i\right)^2/2 + \lambda \|w\|^2/2,
\end{align}
where $\lambda>0$ is a regularization parameter.

\begin{proposition}[Least squares]\label{cor:lse}
Consider the least squares problem \eqref{pbm-lse}.
Assume the step-size $\eta \in (0,\frac{1}{R^{2}+\lambda})$, where $R:=\max_{i}\Vert a_{i}\Vert$ is finite. 
Then, we have the following upper bound:

\begin{equation}
\dimhu \mu_{W|\Sb_n} \leq \frac{\log\left(n/b\right)}{\log(1/(1-\eta\lambda))}.
\label{eqn:prop_lse}
\end{equation}
\end{proposition}
Note that here $\ell$ is only pseudo-Lipschitz $|\ell(w) - \ell(w')|\leq L(1+\|w\|+\|w'\|)\|w-w'\|$, rather than globally Lipschitz. However; the conditions in Proposition~\ref{cor:lse} ensure that $w$ will stay in a bounded region, in which case $\ell$ becomes Lipschitz. Also note that only the logarithm of the Lipschitz constant directly enters the bound.

We observe that, for fixed $n$,the upper bound for  $\dimhu\mu_{W|\Sb_n}$ is decreasing both in $\eta$ and $b$. 
This behavior is not surprising: large $\eta$ results in chaotic behaviors (cf.\ Figures~\ref{fig:cantor},\ref{fig:example}), and in the extreme case where $b=n$, the algorithm becomes deterministic and hence converges to a single point, in which case the Hausdorff dimension becomes $0$. However, the decrement due to $b$ does not automatically grant good generalization performance: since the algorithm becomes deterministic, the stability constant $M$ in \Cref{asmp:decoupling2} can get arbitrarily large, hence the bound in Theorem~\ref{thm:generalization} could become vacuous. This outcome reveals an interesting tradeoff between the Hausdorff dimension and the constant $M$, through the batch-size $b$, and investigating this tradeoff is an interesting future direction. 

We further notice that the numerator in \eqref{eqn:prop_lse} is increasing with $n$, which suggests that the batch-size should be taken in  proportion with $n$ (i.e., setting $m_b$ to a constant value), in order to have a control over $\dimhu \mu_{W|\Sb_n}$. 
Finally, regarding the remaining bounds in this section, even though their forms might differ from \eqref{eqn:prop_lse}, similar remarks also apply. Hence, we will omit the discussion. 

\textbf{Regularized logistic regression. }
Given the data points $z_{i}=(a_{i},y_{i})$, consider regularized logistic regression with the loss:
\begin{equation}\label{pbm-finite-sum}
\ell(w,z_{i}) := \log\left(1+\exp\left(-y_i a_i^T w\right)\right)
+\lambda\|w\|^2/2, 
\end{equation}
where $\lambda>0$ is a regularization parameter. 
We have the following result.

\begin{proposition}[Regularized logistic regression]\label{cor:logistic} 
Consider the regularized logistic regression \eqref{pbm-finite-sum}.
Assume the step-size $\eta<1/\lambda$ and the input data is bounded, 
i.e. $R:=\max_{i}\Vert a_{i}\Vert<2\sqrt{\lambda}$.
We have: 
\begin{equation}
\dimhu \mu_{W|\Sb_n}\leq\frac{\log\left(n/b\right)}{\log(1/(1-\eta\lambda+\frac{1}{4}\eta R^{2}))}.
\end{equation}

\end{proposition}

Next, we consider a non-convex formulation for logistic regression, with $z_{i}=(a_{i},y_{i})$
and the loss
\begin{equation} 
\ell(w,z_{i}) := \rho\left(y_i - \langle w,a_i \rangle \right) + \lambda_r \|w\|^2/2,
 \label{eq-robust-reg-erm}
 \end{equation}
where $\lambda_r>0$ is a regularization parameter and $\rho$ is a non-convex function, where a standard choice is \emph{Tukey's bisquare loss} defined as $\rho_{\text{Tukey}}(t) =1 -( 1 - (t/t_0)^2)^3$ for $|t|\leq t_0$, 
and $\rho_{\text{Tukey}}(t) =1$ for $| t| \geq t_0$, 
and exponential squared loss: $\rho_{\exp}(t) = 1 - e^{-|t|^2/t_0}$, where $t_0>0$ is a tuning parameter.

\begin{proposition}[Non-convex formulation for logistic regression]\label{cor:nonconvex:logistic}
Consider the non-convex formulation for logistic regression \eqref{eq-robust-reg-erm}.
Assume the step-size $\eta<\frac{1}{\lambda_r + R^2(2/t_0)}$, 
where $R = \max_i \|a_i\|<\sqrt{\lambda_{r}t_{0}/2}$.
Then, we have the following upper bound for the Hausdorff dimension:
\begin{equation}
\dimhu \mu_{W|\Sb_n} \leq\frac{\log\left(n/b\right)}{\log(1/(1- \eta \lambda_r + \eta R^2 \frac{2}{t_0}))}.
\end{equation}

\end{proposition}

\textbf{One hidden-layer neural network.} 
Given the data points $z_{i}=(a_{i},y_{i})$.
Let $a_i\in \mathbb{R}^d$ be the input and $y_i \in \mathbb{R}^m$ be the corresponding output, and let $w_r \in \mathbb{R}^{d}$ be the weights of the $r$-th hidden neuron of a one hidden-layer network, and $b_r \in \mathbb{R}$ is the output weight of hidden unit $r$. For simplicity of the presentation, following \cite{ngd-grosse,du2018gradient}, 
we only optimize the weights of the hidden layer, i.e. $w = \begin{bmatrix} w_1^T  w_2^T  \dots  w_m^T \end{bmatrix}$ is the decision variable with the regularized squared loss:
\begin{equation}\label{eqn:one:hidden}
    \ell(w,z_{i}):=\|y_i -\hat{y}_i\|^2 + \lambda\|w\|^2/2, \quad  \hat{y}_i := \sum\limits_{r=1}^m b_r \sigma\left( w_r^T a_i\right),  
\end{equation}
where the non-linearity $\sigma:\mathbb{R}\to\mathbb{R}$ is smooth and $\lambda>0$ is a regularization parameter.

\begin{proposition}[One hidden-layer network]\label{ex:one:hidden:layer} 
Consider the one hidden-layer network \eqref{eqn:one:hidden}.
Assume the step-size $\eta<\frac{1}{2\lambda}$.
Then, we have the following upper bound for the Hausdorff dimension:
\begin{equation}
\dimhu \mu_{W|\Sb_n}\leq\frac{\log\left(n/b\right)}{\log(1/(1-\eta(\lambda-C)))},
\end{equation}
where $C:=M_{y}\|b\|_\infty \|\sigma''\| R^2 +\left( \max_j \|v_j\|_\infty \right)^2<\lambda$, 
where $R:=\max_{i}\Vert a_{i}\Vert$, $M_{y}:=\max_{i}\Vert y_{i}-\hat{y}_{i}\Vert$, and
$v_i:=\left[ 
b_1 \sigma'(w_1^T a_i)a_i \,
b_2 \sigma'(w_2^T a_i)a_i \,
\cdots \\
b_m \sigma'(w_m^T a_i)a_i
\right]^{T}$.
\end{proposition}

\section{Experiments}

\label{sec:experiments}

\begin{figure}[t!]
\centering
\includegraphics[width=1.0\textwidth]{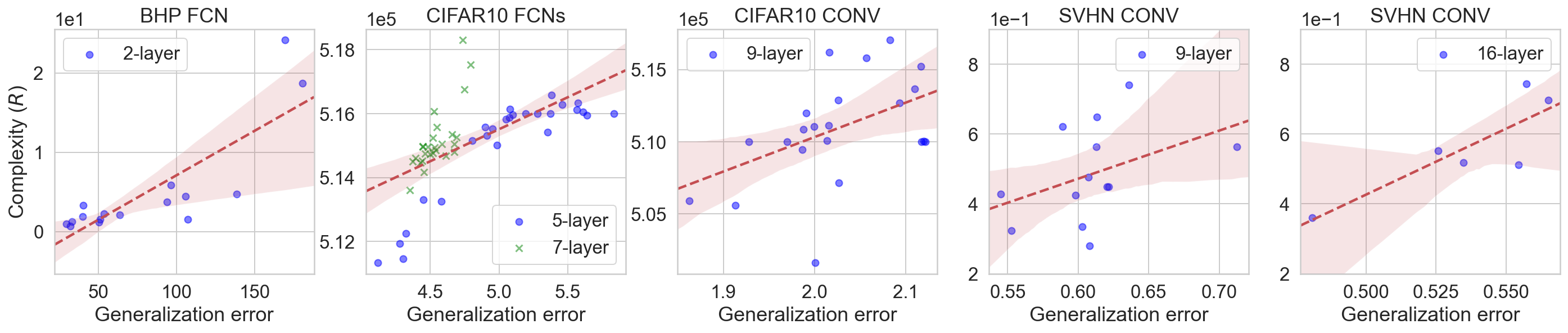}
\vspace{-2mm}
\caption{Estimates of $R$ 
plotted against the generalization error ($|\mathrm{training \ loss} - \mathrm{test \  loss}|$) for VGG11 and FCNs trained on CIFAR10, SVHN and BHP with varying $\eta,b$.
The linear regression of best fit is plotted in red, where shading corresponds to the $95\%$ confidence interval.
For all plots the one-sided p-value, testing whether the null hypothesis that the slope of the line is in-fact negative and not positive, is significantly less than $0.001$, indicating that it is highly likely that $R$ and the generalization error are positively correlated. }
\label{fig:jhi_estimates}
\end{figure}

Our aim now is to empirically demonstrate that the bound in Corollary~\ref{coro-main-thm} is informative, in that it is predictive of a neural network's generalization error. 
As the second term of this bound cannot be evaluated, we focus our efforts on the first term.
Further, because the denominator of the first term is the only term that depends on the invariant measure, 
we want to establish that the inverse of this denominator is predictive of a neural network's generalization error. 
Note however that for complex models, such as modern neural networks, analytically bounding $\|J_{h_i(x)}\|$ becomes highly non-trivial. 

In our experiments, we fix the algorithm to SGD and we develop a numerical method for computing the term \eqref{eqn:newterm}. Noting that $J_{h_{i}}(w) =  I - \eta \nabla^2 \tilde{\Rcal}_i (w)$, for simplicity we denote the inverse of \eqref{eqn:newterm} as the `complexity': 
\[R = 1/\left[\sum\limits_{i=1}^{m_{b}} p_i \int_{\mathbb{R}^d} \log\left(\|J_{h_{i}}(w) \|\right) \rmd\mu_{W|\Sb_n}(w)\right].\] 
To approximate the expectations, we propose the following simple Monte Carlo strategy:
\begin{align}
    R^{-1} \approx \Bigl[1/(N_W N_U)\Bigr]
    \sum\limits_{i=1}^{N_W} \sum\limits_{j=1}^{N_U} \log\left(\|J_{h_{U_j}}(W_i)\|\right),
    \label{eq:jhi_estimate}
\end{align}
where $U_j$ denotes i.i.d.\ random mini-batch indices that are drawn without-replacement from $\{1,\dots,n\}$ 
and $W_i \stackrel{\mathclap{\mbox{\tiny{i.i.d.}}}}{\sim}  \mu_{W|S}$.
Assuming \eqref{assump-mean-contractive} is ergodic \cite{diaconis1999iterated}, we treat the iterates $w_k$ as i.i.d.\ samples from $\mu_{W|\Sb_n}$ for large $k$, hence,  $\log(\|J_{h_{U_j}}(W_i)\|)$ can be computed on these iterates, and \eqref{eq:jhi_estimate} can be computed accordingly. 
Our implementation for computing $\|J_{h_{U_j}}(W_i)\|$ for neural networks with millions of parameters is detailed in the Appendix. Though the size of $J_{h_i}$ is very large in our experiments ($\approx20\text{M}\times20\text{M}$ on average), our algorithm can efficiently compute the norms without constructing $J_{h_{U_j}}(W_i)$, by extending the  approach presented in \cite{pyhessian}.

\begin{figure}
\begin{center}
\includegraphics[width=0.3\textwidth]{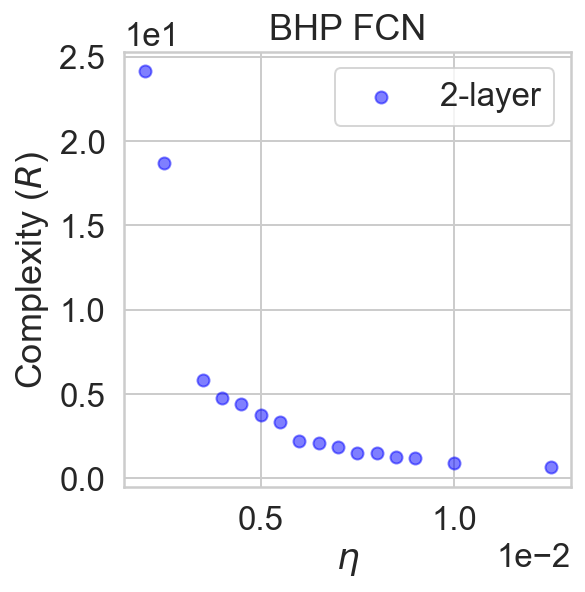}
\end{center}
\caption{Estimated $R$ 
plotted against $\eta$ for 
2 layer FCN trained on BHP.
}
\label{fig:jhi_estimates_lr_br}

\end{figure}

In Figure~\ref{fig:jhi_estimates} we plot the estimates of $R$ for a variety of convolutional (CONV) and fully connected network (FCN) architectures trained on CIFAR10, SVHN and Boston House Prices (BHP). 
For the full details of the models, the hardware used, their run-time, and the convergence criterion used, see Section~\ref{app:network_hyperparams} in the supplement. 
The plot demonstrates that $R$ and generalization error are positively correlated and that this correlation is significant (p-value $\ll 0.001$) for all model architectures.
This provides evidence that the bound on the generalization error in Corollary~\ref{coro-main-thm} is informative. 

To support our findings in 
Section~\ref{sec:examples}
that the bound for the Hausdorff dimension $\dimhu\mu_{W|\Sb_n}$ is monotonically decreasing in the step-size $\eta$, we plot $R$ against $\eta$ in Figure~\ref{fig:jhi_estimates_lr_br} for the networks trained on BHP in Figure~\ref{fig:jhi_estimates}. 
$R$ decreases with increasing $\eta$, clearly backing our findings. 
We note that these results were inconclusive for classification models trained with a cross-entropy loss, in that we could not clearly observe a negative or positive correlation.
Future work will further study this lack of correlation, particular to classification models. 

\vspace{-5pt}
\section{Conclusion}\label{sec:conclusion} 
\vspace{-5pt}

In this work, we investigated stochastic optimization algorithms through the lens of IFSs and studied their generalization properties.
Under some assumptions, 
we showed that the generalization error can be controlled based on the Hausdorff dimension of the invariant measure determined by the iterations, which can lead to tighter bounds than the standard bounds based on the ambient dimension. We proposed an efficient methodology to estimate the Hausdorff dimension in deep learning settings and supported our theory with several experiments on neural networks. We also illustrated our bounds on specific problems and algorithms such as SGD and its preconditioned variants, which unveil new links between generalization, algorithm parameters and the Hessian of the loss.

Our study does not have a direct societal impact as it is largely theoretical. The limitation of our study is its asymptotic nature due to operating on invariant measures. Future work will address obtaining nonasymptotic bounds in terms of the number of iterations $k$. 

\section*{Acknowledgments}

M.G.'s research is supported in part by the grants Office of Naval Research Award Number N00014-21-1-2244, National Science Foundation (NSF) CCF-1814888, NSF DMS-2053485, NSF DMS-1723085. U.\c{S}.'s research is supported by the French government under management of Agence Nationale de la Recherche as part of the ``Investissements d'avenir program, reference ANR-19-P3IA-0001 (PRAIRIE 3IA Institute).
L.Z. is grateful to the support from a Simons Foundation Collaboration Grant and the grant NSF DMS-2053454 from the National Science Foundation.
A. C. was supported by an EPSRC Studentship.

\bibliography{ifs}

\newpage

\appendix

\begin{center}

\Large \bf Appendix

\end{center}

The appendix is organized as follows:

\begin{itemize}
\item Technical background for the proofs.
\begin{itemize}
\item In Section~\ref{sec:dimtheory}, we provide additional background on dimension theory. In particular we define the Minkowski dimension and the local dimension for a \emph{measure}. Then, we provide three existing theoretical results that will be used in our proofs. 
\end{itemize}

\item Additional theoretical results.
\begin{itemize}
\item In Section~\ref{sec:add_sgd}, we provide an upper-bound on the Hausdorff dimension of the invariant measure of SGD, when applied on support vector machines. This result is a continuation of the results given in Section~\ref{sec:examples}.
\item In Section~\ref{sec:precondsgd}, we provide upper-bounds on the Hausdorff dimension of the invariant measure of \emph{preconditioned} SGD on different problems. 
\item In Section~\ref{sec:stochnewton}, we provide an upper-bound on the Hausdorff dimension of the invariant measure of the \emph{stochastic Newton algorithm} applied on linear regression. 
\end{itemize}
\item Details of the experimental results. 
\begin{itemize}
\item In Section~\ref{app:jhi_estimation}, we provide the details of the algorithm that we developed for computing the operator norm $\|I - \eta \nabla^2 \tilde{R}_k(w)\|$ for neural networks. 
\item In Section~\ref{app:network_hyperparams}, we provide the details of the SGD hyperparameters, network architectures, and information regarding hardware/run-time.   
\end{itemize}
\item Proofs.
\begin{itemize}
\item In Section~\ref{sec:proofs}, we provide the proofs all the theoretical results presented in the main document and the Appendix. 
\end{itemize}
\end{itemize}

\section{Further Background on Dimension Theory}
\label{sec:dimtheory}

\subsection{Minkowski dimension of a measure}

Based on the definition of the Minkowski dimension of sets as given in Section~\ref{sec:background}, we can define the Minkowski dimension of a finite Borel measure $\mu$, as follows \cite{pesin2008dimension}:
\begin{align}
    \dimmu \mu :=&\lim_{\delta \rightarrow 0} \inf \left\{\dimmu Z: \mu(Z) \geq 1-\delta\right\}.
\end{align}
Note that in general, we have 
$$\dimmu \mu \leq \inf \left\{\dimmu Z: \mu(Z)=1\right\},$$ 
where the inequality can be strict, see \cite[Chapter 7]{pesin2008dimension}.

\subsection{Local dimensions of a measure} 

It is sometimes more convenient to consider a dimension notion that is defined in a pointwise manner. Let $\mu$ be a finite Borel regular measure on $\mathbb{R}^d$. The lower and upper local (or pointwise) dimensions of $\mu$ at $x \in \mathbb{R}^d$ are respectively defined as follows:
\begin{align}
    \dimll \mu(x) := \liminf_{r \to 0} \frac{\log \mu(B(x,r))}{\log r}, \\
    \dimlu \mu(x) := \limsup_{r \to 0} \frac{\log \mu(B(x,r))}{\log r},
\end{align}
where $B(x,r)$ denotes the ball with radius $r$ about $x$. When the values of these dimensions agree, the common value is called the local (or pointwise) dimension of $\mu$ at $x$, and is denoted by $\diml \mu (x)$. The local dimensions describe the {power-law} behavior of $\mu(B(x,r))$ for small $r$ \cite{alma991001581869705596}. These dimensions are closely linked to the Hausdorff dimension.

\subsection{Existing Results}

The following result upper-bounds the Hausdorff dimension of the invariant measure of an IFS to the constituents of the IFS. We translate the result to our notation. 

\begin{theorem}\cite[Theorem 2.1]{rams2006dimension}\label{thm-rams} 
Consider the IFS \eqref{eq-ifs-2}
and assume that conditions \eqref{assump-lip} and \eqref{assump-mean-contractive} are satisfied and $h_i$ are continuously differentiable with derivatives $h'_i$ that are $\alpha$-H\"older continuous for some $\alpha>0$. The invariant measure $\mu_{W|\Sb_n}$ of the IFS satisfies
$$ \dimhu \mu_{W|\Sb_n} \leq s \quad \mbox{where} \quad s := \frac{\mathrm{Ent}}{\sum_{i=1}^{m_{b}} p_i \int_{x\in\mathbb{R}^d} \log(\|J_{h_{i}}(w)\|) \rmd\mu_{W|\Sb_n}(w) },$$
where $\mathrm{Ent} := \sum_{i=1}^{m_{b}} p_i \log(p_i)$ is the (negative) entropy.
Furthermore, if $h_i$ are conformal and either SOSC or ROSC is satisfied, then we have
$$ \dimh (\mu_{W|\Sb_n})  = \dimhl (\mu_{W|\Sb_n}) = s.$$ 
\end{theorem}

The next two results link the Hausdorff and Minkowski dimensions of a measure to its local dimension. 

\begin{proposition}\cite[Propositions 10.3]{alma991001581869705596}
\label{prop:localdim}
For a finite Borel measure $\mu$, the following identity holds:

\begin{align}
    \dimhu \mu=&\inf \left\{s:  \dimll \mu(x) \leq s \text { for } \mu\text {-almost all } x\right\}.
\end{align}
\end{proposition}

\begin{theorem}\cite[Theorem 7.1]{pesin2008dimension}
\label{thm:mink}
Let $\mu$ be a finite Borel measure on $\mathbb{R}^d$. 
If $\dimlu \mu (x) \leq \alpha$ for $\mu$-almost every $x$, then $\dimmu \mu \leq \alpha$.

\end{theorem}

The next theorem, called Egoroff's theorem, will be used in our proofs repeatedly. It provides a condition for measurable functions to be uniformly continuous in an almost full-measure set. 

\begin{theorem}[Egoroff's Theorem]\cite[Theorem 2.2.1]{bogachev2007measure}
\label{theorem:egoroff}
Let \((X, \mathcal{A}, \mu)\) be a space with a finite nonnegative measure $\mu$ and let $\mu$-measurable functions $f_{n}$ be such that $\mu$-almost everywhere
there is a finite limit $f(x):=\lim_{n \rightarrow \infty} f_{n}(x)$. Then, for every $\varepsilon>0$, there exists
a set $X_{\varepsilon} \in \mathcal{A}$ such that $\mu\left(X \backslash X_{\varepsilon}\right)<\varepsilon$ and the functions $f_{n}$ converge to $f$
uniformly on $X_{\varepsilon}$.
\end{theorem}

\section{Additional Analytical Estimates for SGD}
\label{sec:add_sgd}

\textbf{Support vector machines.} 
Given the data points $z_{i}=(a_{i},y_{i})$ with the input data $a_i$ and the output $y_i \in\{-1,1\}$, consider support vector machines with smooth hinge loss:
\begin{equation}\label{eqn:svm}
\ell(w,z_{i}):=  \ell_\sigma(y_i a_i^T w)  + \lambda \|w\|^2/2,
\end{equation} 
where $\sigma>0$ is a smoothing parameter, $\lambda>0$ is the regularization parameter and $ \ell_\sigma(z) := 1-z + \sigma \log(1 + e^{-\frac{1-z}{\sigma}})$. This loss function is a smooth version of the hinge loss that can be easier to optimize in some settings. In fact, it can be shown that as $\sigma\to 0$, this loss converges to the (non-smooth) hinge loss pointwise.

\begin{proposition}[Support vector machines]\label{ex-svm} 
Consider the support vector machines \eqref{eqn:svm}. 
Assume the step-size $\eta<\frac{1}{\lambda + \|R\|^2/(4\rho)}$, 
where $R := \max_i \|a_i\|$ is finite. 
Then, we have:
\begin{equation}
\dimhu\mu_{W|\Sb_n}\leq\frac{\log\left(n/b\right)}{\log(1/(1-\eta\lambda))}.
\end{equation}

\end{proposition}

\section{Analytical Estimates for Preconditioned SGD}
\label{sec:precondsgd}

We consider the pre-conditioned SGD methods
\begin{align}\label{eq-pre-sgd}
&w_k = w_{k-1} - \eta H^{-1}\nabla\tilde {R}_{k} (w_{k-1}),
\end{align}
for a fixed square matrix $H$. Some choices of $H$ includes a diagonal matrix, a block diagonal matrix or the Fisher-information matrix (see e.g. \cite{ngd-grosse}).
We assume that $H$ is a positive definite matrix, 
and by Cholesky decomposition, we can write $H=SS^{T}$,
where $S$ is a real lower triangular matrix with positive diagonal entries.
If we have $H=JJ^{T}$, where $J$ is the Jacobian,
then the corresponding least square problems is called the Gauss-Newton methods for least squares.
Assume that $H$ 
there exist some $m,M>0$ such that:
\begin{equation}
0\prec mI\preceq H\preceq MI.
\end{equation}

As illustrative examples; in the following, we will consider the setting where we divide $\Sb_n$ into $m_b = n/b$ batches with each batch having $b$ elements, and then we discuss how analytical estimates on the (upper) Hausdorff dimension $\dimhu \mu_{W|\Sb_n}$ can be obtained for some particular problems 
including least squares, regularized logistic regression, support vector machines, and one hidden-layer network.

\textbf{Least squares. }
We consider the least square problem
with data points $z_{i}=(a_{i},y_{i})$ and the loss
\begin{equation}\label{pbm-lse-precond}
\ell(w,z_{i}) :=  \frac{1}{2}\left(a_i^Tw - y_i\right)^2 + \frac{\lambda}{2}\|w\|^2,
\end{equation}
where $\lambda>0$ is a regularization parameter. If we apply preconditioned SGD 
this results in the recursion \eqref{eq-ifs-2} with
\begin{align}
&h_i(w) = M_i w  + q_{i} \quad\text{ with} \quad M_{i}:=I-\eta\lambda H^{-1} - \eta H^{-1}H_{i}, 
\\
&{H}_i := \frac{1}{b}\sum\limits_{j\in S_i } a_j a_j^T, \quad q_i := \frac{\eta}{b}H^{-1} \sum\limits_{j\in S_i } a_j y_j\,,
\nonumber
\end{align}
where $a_j \in \mathbb{R}^d$ are the input vectors, and $y_j$ are the output variables, 
and $\{S_{i}\}_{i=1}^{m_{b}}$ is a partition of $\{1,2,\dots,n\}$ with $|S_i| = b$, where $i=1,2,\ldots,m_{b}$ with $m_{b}=n/b$.
We have the following result.

\begin{proposition}[Least squares]\label{cor:least:squares:2}
Consider \mg{the pre-conditioned SGD method \eqref{eq-pre-sgd}} for the least square problem \eqref{pbm-lse-precond}.
Assume that the step-size $\eta < \frac{m}{R^2 + \lambda}$, where $R:=\max_{i}\Vert a_{i}\Vert$ is finite.
Then, we have the following upper bound for the Hausdorff dimension:
\begin{equation}
\dimhu\mu_{W|\Sb_n}
\leq\frac{\log\left(n/b\right)}{\log(1/(1-\eta M^{-1}\lambda))}.    
\end{equation}

\end{proposition}

\textbf{Regularized logistic regression. }
We consider the regularized logistic regression problem
with the data points $z_{i}=(a_{i},y_{i})$ and the loss:
\begin{equation}\label{logistic:regression:precond}
\ell(w,z_{i}) := \log\left(1+\exp\left(-y_i a_i^T w\right)\right)
+\frac{\lambda}{2}\|w\|^2, 
\end{equation}
where $\lambda>0$ is the regularization parameter.

\begin{proposition}[Regularized logistic regression]\label{cor:logistic:2}
Consider \mg{the pre-conditioned SGD method \eqref{eq-pre-sgd}} for regularized logistic regression \eqref{logistic:regression:precond}.
Assume that the step-size $\eta<m/\lambda$
and $R:=\max_i \|a_i\|<2\sqrt{m\lambda/M}$.
Then, we have the following upper bound for the Hausdorff dimension:
\begin{equation}
\dimhu\mu_{W|\Sb_n}
\leq\frac{\log\left(n/b\right)}{\log(1/(1-\eta M^{-1}\lambda+\frac{1}{4}\eta m^{-1}R^{2}))}.    
\end{equation}

\end{proposition}

Next, we consider a non-convex formulation for logistic regression.
Consider the data points $z_{i}=(a_{i},y_{i})$
and the loss:
\begin{equation}\label{non-convex:logistic:precond}
\ell(w,z_{i}) := \rho\left(y_i - \langle w,a_i \rangle \right) + \frac{\lambda_r}{2} \|w\|^2,
\end{equation}
where $\lambda_r>0$ is a regularization parameter and $\rho$ is a non-convex function. 
We have the following result.

\begin{proposition}[Non-convex formulation for logistic regression]\label{cor:nonconvex:logistic:2}
Consider \mg{the pre-conditioned SGD method \eqref{eq-pre-sgd}} in the non-convex logistic regression setting
\eqref{non-convex:logistic:precond}.
Assume that the step-size $\eta<\frac{m}{\lambda_r + R^2(2/t_0)}$
and $R:=\max_i \|a_i\|<\sqrt{\lambda_{r}t_{0}m/(2M)}$.
Then, we have the following upper bound for the Hausdorff dimension:
\begin{equation}
\dimhu\mu_{W|\Sb_n}\leq\frac{\log\left(n/b\right)}{\log(1/(1- \eta M^{-1} \lambda_r + \eta m^{-1}R^2 \frac{2}{t_0}))}.
\end{equation}
\end{proposition}

\textbf{Support vector machines. } We have the following result \mg{for pre-conditioned SGD when applied to the support vector machines problem \eqref{eqn:svm}}.
\begin{proposition}[Support vector machines]\label{cor:svm:2}
Consider \mg{pre-conditioned SGD  \eqref{eq-pre-sgd}} for \mg{support vector machines \eqref{eqn:svm}}. 
Assume that the step-size $\eta<\frac{m}{\lambda + \|R\|^2/(4\rho)}$
where $R := \max_i \|a_i\|$ is finite.
Then, we have the following upper bound for the Hausdorff dimension:
\begin{equation}
\dimhu\mu_{W|\Sb_n}\leq\frac{\log\left(n/b\right)}{\log(1/(1-\eta M^{-1}\lambda))}.
\end{equation}

\end{proposition}

\textbf{One hidden-layered neural network. }
Consider the one-hidden-layer neural network setting as in Proposition~\ref{ex:one:hidden:layer},  
where the objective is to minimize
the regularized squared loss with the loss function:
\begin{equation}\label{one:hidden:precond}
    \ell(w,z_{i}):=\|y_i -\hat{y}_i\|^2 + \frac{\lambda}{2} \|w\|^2, \quad  \hat{y}_i := \sum_{r=1}^m b_r \sigma\left( w_r^T a_i\right),  
\end{equation}
where the non-linearity $\sigma:\mathbb{R}\to\mathbb{R}$ is smooth and $\lambda>0$ is the regularization parameter. 

\begin{proposition}[One hidden-layer network]\label{cor:one:hidden:2}
Consider the one-hidden-layer network \eqref{one:hidden:precond}.
Assume that $\eta<\frac{m}{C+\lambda}$ and $\lambda>\frac{M}{m}C$, 
where $C$ is defined in Corollary~\ref{ex:one:hidden:layer}. 
Then, we have the following upper bound for the Hausdorff dimension:
\begin{equation}
\dimhu\mu_{W|\Sb_n}\leq\frac{\log\left(n/b\right)}{\log(1/(1-\eta(M^{-1}\lambda-m^{-1}C)))}.
\end{equation}

\end{proposition}

\section{Analytical Estimates for Stochastic Newton}
\label{sec:stochnewton}

We consider the stochastic Newton method
\begin{align}\label{eqn:stoc-newton}
&w_k = w_{k-1} - \eta [\tilde{H}_k(w_{k-1})]^{-1}\nabla R_{k} (w_{k-1}), \quad \mbox{where} \quad  \tilde{H}_k(w) := (1/b) \sum\limits_{i\in \Omega_k} \nabla^2 \ell(w,z_{i}), 
\nonumber
\end{align}
see e.g. \cite{roosta2016sub}, where $\Omega_{k}=S_{i}$ with probability $p_{i}$ with $i=1,2,\ldots,m_{b}$, where $m_{b}=n/b$.

For simplicity, we focus on the least square problem, 
with the data points $z_{i}=(a_{i},y_{i})$ and the loss:
\begin{equation}\label{pbm-lse-stoch-Newton}
\ell(w,z_{i}):=\frac{1}{2}\left(a_i^Tw - y_i\right)^2 + \frac{\lambda}{2}\|w\|^2,
\end{equation}
where $\lambda>0$ is a regularization parameter. 
If we apply stochastic Newton 
this results in the recursion \eqref{eq-ifs-2} with
\begin{align}
&h_i(w) = M_i w  + q_{i} \quad\text{ with} \quad M_{i}:=(1-\eta)I, 
\\
&\tilde{H}_i := \frac{1}{b}\sum\limits_{j\in S_i } a_j a_j^T+\lambda I, \quad q_i := \frac{\eta}{b}\tilde{H}_{i}^{-1} \sum\limits_{j\in S_i } a_j y_j\,,
\nonumber
\end{align}
where $a_j\in \mathbb{R}^d$ are the input vector, and $y_j$ are the output variable, 
and $\{S_{i}\}_{i=1}^{m_{b}}$ is a partition of $\{1,2,\dots,n\}$ with $|S_i| = b$, where $i=1,2,\ldots,m_{b}$ with $m_{b}=n/b$.
Therefore, $J_{h_{i}}(w)=(1-\eta)I$.
By following the similar argument as in the proof of Proposition~\ref{cor:least:squares:2}, we conclude that for any $\eta\in(0,1)$,
\begin{equation}
\dimhu\mu_{W|\Sb_n}\leq\frac{\log\left(n/b\right)}{\log(1/(1-\eta))},
\end{equation}

where the upper bound is decreasing in step-size $\eta$ and batch-size $b$.

\section{Estimating the Complexity $R$ for SGD}
\label{app:jhi_estimation}

Estimating $R$, as detailed in Equation~\eqref{eq:jhi_estimate}, requires drawing $N_W$ samples from the invariant measure and $N_U$ batches of from the training data. 
As mentioned in the main text, to approximate the summation over $N_W$ samples from the invariant measure, assuming \eqref{assump-mean-contractive} is ergodic \cite{diaconis1999iterated}, we treat the iterates $w_k$ as i.i.d.\ samples from $\mu_{W|S}$ for large $k$, hence, the norm of the Jacobian $\log(\|J_{h_{I_j}}(W_i)\|)$ can be efficiently computed on these iterates.
Thus, we first we train a neural-network to convergence, whereby convergence is defined as the model reaching some accuracy level (if the dataset is a classification task) \textit{and} achieving a loss below some threshold on training data. 
We assume that after convergence the SGD iterates will be drawn from the invariant measure. 
As such we run the training algorithm for another 200 iterates, saving a snapshot of the model parameters at each step, such that $N_W = 200$ in Equation~\eqref{eq:jhi_estimate}. 
For each of these snapshots we estimate the spectral norm $\|J_{h_{I}}(W)\|$ using a simple modification of the power iteration algorithm of \cite{pyhessian}, detailed in Section~\ref{app:power_iter_mod} below. 
This modified algorithm is scalable to neural networks with millions of parameters and we apply it to 50 of the batches used during training, such that $N_U = 50$ in \eqref{eq:jhi_estimate}.

\subsection{Power Iteration Algorithm for $\|J_{h_{i}}(w)\|$}
\label{app:power_iter_mod}
We re-purpose the power iteration algorithm of \cite{pyhessian} adding a small modification that allows for the estimation of the spectral norm $\|J_{h_{i}}(w)\|$. 
We first note that 
\begin{equation}
    J_{h_{i}}(w) =  I - \eta \nabla^2 \tilde{\Rcal}_i (w)
\end{equation}
where $\nabla^2 \tilde{\Rcal}_i (w)$ is the Hessian for the $i^{th}$ batch. 
As such our power iteration algorithm needs to estimate the operator norm of the matrix $I - \eta \nabla^2 \tilde{\Rcal}_i (w)$ and not just that of the Hessian of the network.  
To do this we just need to change the `vector-product' step of the power-iteration algorithm of  \cite{pyhessian}.
Our modified method has the same convergence guarantees, namely that the method will converge to the `true' top eigenvalue if this eigenvalue is `dominant', in that it dominates all other eigenvalues in absolute value, i.e if $\lambda_1$ is the top eigenvalue then we must have that: 
\[
|\lambda_1|>|\lambda_2|\geq \dots |\lambda_n|
\]
to guarantee convergence.

\begin{algorithm}[h] 
\DontPrintSemicolon
\caption{Power Iteration for Top Eigenvalue Computation of $J_{h_{i}}(w)$}
\label{alg:power_iteration}
    \SetAlgoLined
    \KwInput{Network Parameters: $w$, Loss function: $f$,
    Learning rate: $\eta$}
    Compute the gradient of $\theta$ by backpropagation, i.e., compute $g_w=\frac{d f}{d w}$.
    
    Draw a random vector $v$ from $N(0,1)$  (same dimension as $w$).
    
    Normalize $v$, $v=\frac{v}{\|v\|_2}$
    
    \For(\ \ \quad \quad\quad\quad\quad\tcp*[h]{Power Iteration}){i $=1,2,\ldots$}{
        Compute $gv = g_\theta^Tv$ \tcp*{Inner product}
        
        Compute $Hv$ by backpropagation, $Hv = \frac{d(gv)}{dw}$ \tcp*{Get Hessian vector product}  
         Compute $J_{h_{i}}v$, $J_{h_{i}}v = (I-\eta H)v = v - \eta Hv$  \tcp*{Get $J_{h_{i}}$ vector product} 
        Normalize and reset $v$, $v = \frac{J_{h_{i}} v}{\|J_{h_{i}} v\|_2}$
    }
\end{algorithm}

\section{Experiment Hyperparameters}
\label{app:network_hyperparams}

\paragraph{Training Parameters:}
All models in Figures~\ref{fig:jhi_estimates} and~\ref{fig:jhi_estimates_lr_br} were trained using SGD with batch sizes of $50$ or $100$ and were considered to have converged for CIFAR10 and SVHN if they reached $100\%$ accuracy and less than $0.0005$ loss on the training set. 
For BHP convergence was considered to have been achieved after $100000$ training steps. 
For all models except VGG16 in Figures~\ref{fig:jhi_estimates} and~\ref{fig:jhi_estimates_lr_br} we use learning rates in 
\begin{align*}
    \big\{0.0075, 0.02, 0.025, 0.03, 0.04, 0.06, 0.07, 0.075, 0.08, 0.09, 0.1, 0.11,  0.12, \\ 0.13,  0.14, 0.15, 0.16,  
    0.17, 0.18, 0.19, 0.194, 0.2, 0.22, 0.24, 0.25, 0.26\big\}.
\end{align*}
VGG16 models were trained with learning rates in $\{0.0075, 0.02, 0.03, 0.06, 0.07, 0.08\}$.

\paragraph{Network Architectures:}
BHP FCN had 2 hidden layers and were 10 neurons wide. Similarly CIFAR10 FCN were 5 and 7 layers deep with 2048 neurons per layer. 
9-layer CONV networks were VGG11 networks with the final 2 layers removed. 
16-layer CONV networks were simply the standard implementation of VGG16 networks.

\paragraph{Run-time:} 
The full battery of fully connected models split over two \textit{GeForce GTX 1080} GPUs took two days to train to convergence and the subsequent power iterations took less than a day. 
Similarly the full gamut of VGG11 models took a day to train to convergence over four \textit{GeForce GTX 1080} GPUs and the subsequent power iterations took less than a day to converge. 
The VGG16 models took a day to train over four \textit{GeForce GTX 1080} GPUs but the power iterations \textbf{for each model} took roughly 24 hours on a single \textit{GeForce GTX 1080} GPU.

\section{Postponed Proofs}
\label{sec:proofs}

\subsection{Proof of Proposition~\ref{lem:existence}}

\begin{proof}
Denote $\alpha := \dimhu \mu_{W|\Sb_n}$. By Assumption \Cref{asmp:localreg}, we have $$\dimll \mu_{W|\Sb_n}(w) = \dimlu \mu_{W|\Sb_n}(w),$$ for $\mu_{W|\Sb_n}$-a.e.\ $w$, and by Proposition~\ref{prop:localdim} we have
\begin{align}
    \dimlu \mu_{W|\Sb_n}(w) \leq \alpha + \epsilon,
\end{align}
for all $\epsilon >0$ and for $\mu_{W|\Sb_n}$-a.e.\ $w$. By invoking Theorem~\ref{thm:mink}, we obtain
\begin{align}
    \dimmu \mu_{W|\Sb_n} \leq \alpha + \epsilon.
\end{align}
Since this holds for any $\epsilon$, $\dimmu \mu_{W|\Sb_n} \leq \alpha$. 

By definition, we have for almost all $\Sb_n$:
\begin{align}
\dimmu \mu_{W|\Sb_n} = \lim_{\delta \rightarrow 0} \inf \left\{\dimmu A: \mu_{W|\Sb_n}(A) \geq 1-\delta\right\}.
\end{align}
Hence, given a sequence $(\delta_k)_{k\geq 1}$ such that $\delta_k \downarrow 0$, and $\Sb_n$, and any $\epsilon >0$, there is a $k_0=k_0(\epsilon)$ such that $k \geq k_0$ implies 
\begin{align}
\inf \left\{ \dimmu A : \mu_{W|\Sb_n}(A) \geq 1-\delta_k \right\}  \leq&  \dimmu \mu_{W|\Sb_n} +\epsilon \\ \leq& \alpha + \epsilon.    
\end{align}
Hence, for any $\epsilon_1 >0$ and $k \geq k_0$, we can find a bounded Borel set $A_{\Sb_n,k}$, such that $\mu_{W|\Sb_n}(A_{\Sb_n,k})\geq 1-\delta_k$, and 
\begin{align}
    \dimmu A_{\Sb_n,k} \leq \alpha + \epsilon + \epsilon_1.
\end{align}
Note that the boundedness of $A_{\Sb_n,k}$ follows from the fact that its upper-Minkowski dimension is finite. By choosing $\epsilon = \epsilon_1 = \frac{\veps}{2}$, it yields the desired result. This completes the proof.
\end{proof}

\subsection{Proof of Theorem~\ref{thm:generalization}}

\begin{proof}
We begin similarly to the proof of Proposition~\ref{lem:existence}.

Denote 
$$\alpha(\Sb,n) := \dimhu \mu_{W|\Sb_n}.$$ 
By Assumption \Cref{asmp:localreg}, we have $\dimll \mu_{W|\Sb_n}(w) = \dimlu \mu_{W|\Sb_n}(w)$ for $\mu_{W|\Sb_n}$-almost every $w$, and by Proposition~\ref{prop:localdim} we have
\begin{align}
    \dimlu \mu_{W|\Sb_n}(w) \leq \alpha(\Sb,n) + \epsilon,
\end{align}
for all $\epsilon >0$ and for $\mu_{W|\Sb_n}$-a.e.\ $w$. By invoking Theorem~\ref{thm:mink}, we obtain
\begin{align}
    \dimmu \mu_{W|\Sb_n} \leq \alpha(\Sb,n) + \epsilon.
\end{align}
Since this holds for any $\epsilon >0$, $\dimmu \mu_{W|\Sb_n} \leq \alpha(\Sb,n)$.

By definition, we have for all $\Sb$ and $n$:
\begin{align}
\dimmu \mu_{W|\Sb_n} = \lim_{\delta \to 0} \inf \left\{\dimmu A: \mu_{W|\Sb_n}(A) \geq 1-\delta\right\}.
\end{align}
Hence, for each $n$, there exists a set $\Omega_n$ of full measure such that 
\begin{align}
\label{eqn:proof_interm_omega}
f^n_\delta(\mathbf{S}) := \inf \left\{\dimmu A: \mu_{W|\Sb_n}(A) \geq 1-\delta\right\} \to \dimmu \mu_{W|\Sb_n},
\end{align}
for all $\mathbf{S} \in \Omega_n$. Let $\Omega^\ast := \cap_n \Omega_n$. Then for $\mathbf{S}\in \Omega^\ast$ we have that for all $n$
\begin{align}
f^n_\delta(\mathbf{S}) \to  \dimmu \mu_{W|\Sb_n},
\end{align}
and therefore, on this set we also have
$$\sup_n \frac{1}{\xi_n}\min\left\{1,  \left|f^n_\delta(\mathbf{S})- \dimmu \mu_{W|\Sb_n}\right|\right\} \to 0,$$
where $\xi_n$ is a monotone increasing sequence such that $\xi_n \geq 1$ and $\xi_n \to \infty$. 

By applying Theorem~\ref{theorem:egoroff} to the collection of random variables:
\begin{align}
F_\delta(\Sb) := \sup_n \frac{1}{\xi_n}\min\left\{1,  \left|f^n_\delta(\mathbf{S})- \dimmu \mu_{W|\Sb_n}\right|\right\},
\end{align}
for any $\zeta>0$, we can find a subset $\mathfrak{Z}\subset \Zcal^\infty$, with probability at least $1-\zeta$ under $\pi^\infty$, such that on $\mathfrak{Z}$ the convergence is uniform, that is 
\begin{align}
\sup_{\mathbf{S}\in \mathfrak{Z}}\sup_n \frac{1}{\xi_n}\min\left\{1,  \left|f^n_\delta(\mathbf{S})- \dimmu \mu_{W|\Sb_n}\right|\right\} \leq c(\delta),
\end{align}
where for any $\zeta$, $c(\delta) := c(\delta; \zeta) \to 0$ as $\delta \to 0$.

Hence, for any $\delta$, $\Sb \in\mathfrak{Z}$, and $n$, we have
\begin{align}
f^n_{\delta}(\Sb)   \leq&  \dimmu \mu_{W|\Sb_n} + \xi_n c(\delta)  \\
\leq& \alpha(\Sb,n) + \xi_n c(\delta).
\end{align}
Consider a sequence $(\delta_k)_{k\geq 1}$ such that $\delta_k \downarrow 0$. 
Then, for any $\Sb \in \mathfrak{Z}$ and $\epsilon >0$, we can find a bounded Borel set $A_{\Sb_n,k}$, such that $\mu_{W|\Sb_n}(A_{\Sb_n,k})\geq 1-\delta_k$, and 
\begin{align}
    \label{eqn:thm_interm_dimm}
    \dimmu A_{\Sb_n,k} \leq \alpha(\Sb,n) + \xi_n c(\delta_k)+ \epsilon.
\end{align}
Define the set
\begin{align}
    \Wcal_{n,\delta_k} := \bigcup_{\Sb\in \Zcal^\infty} A_{\Sb_n,k}.
\end{align}
By using $\Gcal(w) :=|\Rcal(w) - \hat{\Rcal}(\wb,\Sb_n)|$, under the joint distribution of $(W,\Sb_n)$, such that $\Sb \sim \pi^\infty$ and $W \sim \mu_{W|\Sb_n}$, we have: 
\begin{align}
    \mathbb{P} \left( \Gcal(W) > \veps \right) \leq& \zeta +  \mathbb{P} \left( \{ \Gcal(W) > \veps \}  \cap \{\Sb \in \mathfrak{Z}\} \right)  \\
    \leq& \zeta + \delta_k + \mathbb{P} \left( \{ \Gcal(W) > \veps \} \cap \{W \in A_{\Sb_n,k}\} \cap \{\Sb \in \mathfrak{Z}\} \right) \\
    \leq & \zeta + \delta_k + \mathbb{P}\left( \left\{\sup_{w \in A_{\Sb_n,k}} \Gcal(w) > \veps\right\}   \cap \{\Sb \in \mathfrak{Z}\} \right). \label{eqn:proof_intermbound}
\end{align}

Now, let us focus on the last term of the above equation. 

First we observe that as $\ell$ is $L$-Lipschitz, so are $\Rcal$ and $\hat{\Rcal}$. Hence, by considering the particular forms of the $\beta$-covers in \Cref{asmp:decoupling2}, for any $\wb' \in \rset^d$ we have:
  \begin{align}
    \Gcal(w) \leq \Gcal(w')+2 L\left\|\wb-\wb^{\prime}\right\|,
  \end{align}
which implies
\begin{align}
 \label{eqn:interm_lip}
    \sup_{\wb \in A_{\Sb_n,k}} \mathcal{G}(w) \leq  \max_{\wb \in N_{\beta_n}(A_{\Sb_n,k})} \mathcal{G}(w) +2 L\beta_n.
\end{align}

Now, notice that the $\beta$-covers of \Cref{asmp:decoupling2} still yield the same Minkowski dimension in \eqref{eqn:dimmink} \cite{simsekli2020hausdorff}. Then by definition, we have for all $\Sb$ and $n$:
\begin{align}
\limsup_{\beta \to 0} \frac{\log |N_\beta(A_{\Sb_n,k})|}{\log(1/\beta)} = \lim_{\beta\to 0} 
\sup_{r<\beta}\frac{\log |N_r(A_{\Sb_n,k})|}{\log(1/r)} = \dimmu A_{\Sb_n,k} := \dm(\Sb,n,k).
\end{align}
Hence for each $n$
\begin{align}
g^{n,k}_\beta(\mathbf{S}):= \sup_{\mathbb{Q}\ni r<\delta}\frac{\log |N_r(A_{\Sb_n,k})|}{\log(1/r)} \to \dm(\mathbf{S},n,k),\label{eq:limitdefnofdh}
\end{align}
almost surely. By using the same reasoning in \eqref{eqn:proof_interm_omega}, we have, for each $n$, there exists a set $\Omega'_n$ of full measure such that 
\begin{align}
g^{n,k}_\beta(\mathbf{S}) = \sup_{\mathbb{Q}\ni r<\beta}\frac{\log |N_r(A_{\Sb_n,k})|}{\log(1/r)} \to \dm(\mathbf{S},n,k),
\end{align}
for all $\Sb \in \Omega'_n$. Define $\Omega^{\ast\ast} := (\cap_n \Omega'_n) \cap \Omega^\ast$. Hence, on $\Omega^{\ast\ast}$ we have: 
\begin{align}
G^k_\beta(\Sb) := \sup_n \frac{1}{\xi_n}\min\left\{1,  \left|g^{n,k}_\beta(\mathbf{S})- \dm(\Sb,n,k)\right|\right\} \to 0,
\end{align}

By applying Theorem~\ref{theorem:egoroff} to the collection $\{G^k_\beta(\Sb)\}_\beta$, for any $\zeta_1>0$ we can find a subset $\mathfrak{Z}_1\subset \Zcal^\infty$, with probability at least $1-\zeta_1$ under $\pi^\infty$, such that on $\mathfrak{Z}_1$ the convergence is uniform, that is 
\begin{align}
\sup_{\mathbf{S}\in \mathfrak{Z}_1}\sup_n \frac{1}{\xi_n}\min\{1,  |g^{n,k}_\beta(\mathbf{S})- \dm(\Sb,n,k)|\} \leq c'(\beta),
\end{align}
where for any $\zeta_1$, $c'(\beta) := c'(\beta; \zeta_1, \delta_k) \to 0$ as $\beta \to 0$.

Hence, denoting $\mathfrak{Z}^\ast := \mathfrak{Z} \cap \mathfrak{Z}_1$ by using \eqref{eqn:thm_interm_dimm} we have:
\begin{align*}
\{\mathbf{S}\in \mathfrak{Z}^\ast \} \subseteq 
\bigcap_{n}\left\{ \left|N_{\beta}\left(A_{\mathbf{S}_n,k}\right)\right| \leq \left(\frac{1}{\beta}\right)^{\alpha(\mathbf{S},n)+\xi_n c(\delta_k)  +\xi_n c'(\beta) + \epsilon}\right\}.
\end{align*}

Let $(\beta_n)_{n \geq 0}$ be a decreasing sequence such that $\beta_n \in \mathbb{Q}$ for all $n$ and $\beta_n\to 0$. 
We then have

\begin{align*}
 \pr \Biggl(\{\Sb \in \mathfrak{Z}\}\cap & \left\{\max_{\wb \in N_{\beta_n}(A_{\Sb_n,k})} \Gcal_n(\wb) \geq \veps\right\}\Biggr) \\
&\leq  \pr \Biggl(\{\Sb \in \mathfrak{Z}^\ast\}\cap \Big\{\max_{\wb \in N_{\beta_n}(A_{\textbf{S}_n,k})} \Gcal_n(\wb) \geq \veps\Big\}\Biggr) +\zeta_2.
\end{align*}

For $\rho>0$ and $m\in \mathbb{N}_+$ let us define the interval $J_m(\rho):=(m\rho, (m+1)\rho]$. Furthermore, for any $t>0$ define 
\begin{align}
\veps(t):= \sqrt{\frac{2\nu^2}{n} \Big[\log(1/\beta_n)\left(t + \xi_n c(\delta_k) + \xi_n c'(\beta_n) + \epsilon \right)+ \log(M/\zeta_2) \Big] }.
\end{align}

For notational simplicity, denote $N_{\beta_n,k} := N_{\beta_n}(\Wcal_{n,\delta_k}) $ and
\begin{align}
\tilde{\alpha}(\Sb,n,k,\epsilon) := \alpha(\mathbf{S},n)+\xi_n c(\delta_k) + \xi_n c'(\beta_n) + \epsilon.
\end{align}

Let $d^\ast$ be the smallest real number such that $\alpha(\Sb,n) \leq d^\ast$ almost surely\footnote{Notice that we trivially have $d^\ast\leq d$; yet, $d^\ast$ can be much smaller than $d$.}, we therefore have:

\begin{align*}
&\pr \Biggl( \{\Sb \in \mathfrak{Z}\}\cap \left\{\max_{\wb \in N_{\beta_n}(A_{\textbf{S}_n,k})} \Gcal_n(\wb) \geq \veps(\alpha(\mathbf{S},n))\right\}\Biggr)\\
&\leq  \zeta_2 + \pr \Biggl(\left\{ \left|N_{\beta_n}(A_{\mathbf{S}_n,k})\right|\leq \left(\frac1{\beta_n}\right)^{\tilde{\alpha}(\Sb,n,k,\epsilon)} \right\}
\\
&\qquad\qquad\qquad\qquad\qquad\cap \Big\{\max_{\wb \in N_{\beta_n}(A_{\mathbf{S}_n},k)} |\hat{\Rcal}_n(\wb)-\Rcal(\wb)| \geq \veps(\alpha(\mathbf{S},n))\Big\}\Biggr) \\
&= \zeta_2+ 
\sum_{m=0}^{\lceil \frac{d^\ast}{\rho} \rceil} 
\pr \Biggl(\left\{ \left|N_{\beta_n}(A_{\mathbf{S}_n,k})\right|\leq \left(\frac1{\beta_n}\right)^{\tilde{\alpha}(\Sb,n,k,\epsilon)} \right\}\\
&\qquad \qquad \qquad\qquad \cap \Big\{\max_{\wb \in N_{\beta_n}(A_{\mathbf{S}_n,k})} \Gcal_n(\wb) \geq \veps(\alpha(\mathbf{S},n))\Big\}
\cap \left\{\alpha(\mathbf{S},n) \in J_m(\rho) \right\} 
\Biggr)
\\
&= \zeta_2+ 
\sum_{m=0}^{\lceil \frac{d^\ast}{\rho} \rceil }  
\pr \Biggl(\left\{ \left|N_{\beta_n}(A_{\mathbf{S}_n,k})\right|\leq \left(\frac1{\beta_n}\right)^{\tilde{\alpha}(\Sb,n,k,\epsilon)} \right\}\cap \left\{\alpha(\mathbf{S},n) \in J_m(\rho) \right\} \\
&\qquad \qquad \qquad \cap \bigcup_{w\in N(\beta_n)} 
\bigg(\left\{\wb \in N_{\beta_n}(A_{\mathbf{S}_n,k})\right\}\cap\left\{ \Gcal_n(\wb) \geq \veps(\alpha(\mathbf{S},n)) \right\} \bigg)
\Biggr)
\\
&\leq \zeta_2+ \sum_{m=0}^{\lceil \frac{d^\ast}{\rho} \rceil } \sum_{w\in N_{\beta_{n},k}}\pr \Bigg(  \Big\{\Gcal_n(\wb) \geq \veps(m\rho)\Big\} \\
&\qquad \cap
\Big\{w\in N_{\delta_{n}}(A_{\mathbf{S}_n,k})\Big\} \cap \left\{ \left|N_{\beta_n}(A_{\mathbf{S}_n,k})\right|\leq \left(\frac1{\beta_n}\right)^{\tilde{\alpha}(\Sb,n,k,\epsilon)} \right\}\cap \left\{\alpha(\mathbf{S},n) \in J_m(\rho) \right\}\Bigg),
\end{align*}
where we used the fact that on the event $\alpha(\mathbf{S},n) \in J_m(\rho)$, $\veps(\alpha(\mathbf{S},n))\geq \veps(m\rho)$.

Notice that the events
$$\Big\{w\in N_{\beta_n}(A_{\mathbf{S}_n,k})\Big\}, \left\{|N_{\beta_n}(A_{\mathbf{S}_n,k})|\leq (1/\beta_n)^{\tilde{\alpha}(\Sb,n,k,\epsilon)} \right\}, \left\{\alpha(\mathbf{S},n) \in J_m(\rho) \right\}$$ 
are in  $\mathfrak{G}$.

On the other hand, the event $\{\Gcal_n(\wb) \geq \veps(m\rho)\}$ is clearly in $\mathfrak{F}$ (see \Cref{asmp:decoupling2} for definitions).

Therefore, we have

\begin{align*}
&\pr \Biggl( \{\Sb \in \mathfrak{Z}\}\cap \left\{ \max_{\wb \in N_{\beta_n}(A_{\mathbf{S}_n,k})} \Gcal_n(\wb) \geq \veps(\alpha(\mathbf{S},n)) \right\}\Biggr)\\ 
&\leq \zeta_2+ M\sum_{m=0}^{\lceil \frac{d^\ast}{\rho} \rceil }\sum_{w\in N_{\beta_n,k}}\pr \left( \Big\{\Gcal_n(\wb) \geq \veps(m\rho)\Big\}\right) \\
&\qquad \times \pr \Biggl(
\Big\{w\in N_{\beta_n}(A_{\mathbf{S}_n,k})\Big\} \cap \left\{ \left|N_{\beta_n}(A_{\mathbf{S}_n,k})\right|\leq \left(\frac1{\beta_n}\right)^{\tilde{\alpha}(\Sb,n,k,\epsilon)} \right\}\cap \left\{\alpha(\mathbf{S},n) \in J_m(\rho) \right\}\Bigg),
\end{align*}

Recall that 
$\mathcal{G}_n(w)=\frac{1}{n}\sum_{i=1}^n [\ell(w,z_i)-\mathbb{E}_{z\sim\pi} \ell(w,z)]$. Since the $(z_i)_i$ are i.i.d.\ by Assumption~\ref{asmp:subexp} it follows that 
$\mathcal{G}_n(w)$ is $(\nu/\sqrt{n},\kappa/n)$-sub-exponential and from \cite[Proposition~2.9]{wainwright2019high} we have that 
$$\pr \left( \Big\{\Gcal_n(\wb) \geq \veps(m\rho)\Big\}\right)
\leq 2 \exp\left(-\frac{n\veps(m\rho)^2}{2\nu^2} \right),$$
as long as $\veps(m\rho)\leq \nu^2/\kappa$.

For $n$ large enough we may assume that $\veps(d^\ast) \leq \nu^2/\kappa$, and thus

\begin{align*}
&\pr \Biggl( \{\Sb \in \mathfrak{Z}\}\cap \left\{ \max_{\wb \in N_{\beta_n}(A_{\mathbf{S}_n,k})} \Gcal_n(\wb) \geq \veps(\alpha(\mathbf{S},n)) \right\}\Biggr)\\ 
&\leq  2M\sum_{m=0}^{\lceil \frac{d}{\rho} \rceil }\re^{-\frac{2n\veps^2(m\rho)}{B^2}}
\sum_{w\in N_{\beta_{n},k}}\pr \Biggl(
\Big\{w\in N_{\beta_n}(A_{\mathbf{S}_n,k})\Big\} \cap \left\{ \left|N_{\beta_n}(A_{\mathbf{S}_n,k})\right|\leq \left(\frac1{\beta_n}\right)^{\tilde{\alpha}(\Sb,n,k,\epsilon)} \right\} \\
&\qquad \qquad \qquad \qquad \qquad \qquad\qquad \qquad \cap \left\{\alpha(\mathbf{S},n) \in J_m(\rho) \right\}\Bigg)+\zeta_2
\\
&\leq 2M\sum_{m=0}^{\lceil \frac{d}{\rho} \rceil }\re^{-\frac{n\veps^2(m\rho)}{2\nu^2}}
\sum_{w\in N_{\beta_n,k}}\mathbb{E} \Biggl[
\mathds{1}\Big\{w\in N_{\beta_n}(A_{\mathbf{S}_n,k})\Big\}  \\
&\qquad \qquad \qquad 
\times  \mathds{1}\left\{ \left|N_{\beta_n}(A_{\mathbf{S}_n,k})\right|\leq \left(\frac1{\beta_n}\right)^{\tilde{\alpha}(\Sb,n,k,\epsilon)} \right\}\times \mathds{1}\left\{\alpha(\mathbf{S},n) \in J_m(\rho) \right\}\Bigg]
+\zeta_2
\\
&\leq 2M\sum_{m=0}^{\lceil \frac{d}{\rho} \rceil }\re^{-\frac{n\veps^2(m\rho)}{2\nu^2}}
\mathbb{E} \Biggl[\sum_{w\in N_{\beta_n,k}}
\mathds{1}\Big\{w\in N_{\beta_n}(A_{\mathbf{S}_n,k})\Big\}   \\
&\qquad \qquad \qquad 
\times  \mathds{1}\left\{ \left|N_{\beta_n}(A_{\mathbf{S}_n,k})\right|\leq \left(\frac1{\beta_n}\right)^{\tilde{\alpha}(\Sb,n,k,\epsilon)} \right\}
\times \mathds{1}\left\{\alpha(\mathbf{S},n) \in J_m(\rho) \right\}\Bigg]
+\zeta_2
\numberthis \label{eqn:fubini}\\
&\leq 2M\sum_{m=0}^{\lceil \frac{d}{\rho} \rceil }\re^{-\frac{n\veps^2(m\rho)}{2\nu^2}}
\mathbb{E} \Biggl[ |N_{\beta_n}(A_{\mathbf{S}_n,k})| \times  \mathds{1}\left\{ \left|N_{\beta_n}(A_{\mathbf{S}_n,k})\right|\leq \left(\frac1{\beta_n}\right)^{\tilde{\alpha}(\Sb,n,k,\epsilon)} \right\} \\
&\qquad \qquad \qquad \qquad \qquad \qquad\qquad \qquad \times \mathds{1}\left\{\alpha(\mathbf{S},n) \in J_m(\rho) \right\}\Bigg]
+\zeta_2\\
&= \zeta_2+ 2M\sum_{m=0}^{\lceil \frac{d}{\rho} \rceil }\re^{-\frac{n\veps^2(m\rho)}{2\nu^2}}
\mathbb{E} \left[\left[\frac{1}{\beta_n} \right]^{\tilde{\alpha}(\Sb,n,k,\epsilon)}   \times \mathds{1}\left\{\alpha(\mathbf{S},n) \in J_m(\rho) \right\}\right],
\end{align*}
where \eqref{eqn:fubini} follows from Fubini's theorem.

Now, notice that the mapping $t \mapsto \veps^2(t)$ is linear with derivative bounded by 
$$\frac{2\nu^2}{n} \log(1/\beta_n).$$
Therefore, on the event $\{ \alpha(\mathbf{S},n)\in J_m(\rho)\}$ we have 
\begin{align}
\veps^2(\alpha(\mathbf{S},n)) - \veps^2(m\rho) \leq& (\alpha(\mathbf{S},n) - m\rho) \frac{2\nu^2}{n} \log(1/\beta_n)\\
\leq&  \rho \frac{2\nu^2}{n} \log(1/\beta_n).
\end{align}
By choosing $\rho = \rho_n =1/\log(1/\beta_n)$, we have
$$
\veps^2(m\rho_n) \geq \veps^2\left( \alpha(\mathbf{S},n)\right) - \frac{2\nu^2}{n}.
$$
Therefore, we have
\begin{align*}
 &\pr \Biggl( \{\Sb \in \mathfrak{Z}\}\cap \Bigl\{ \max_{\wb \in N_{\beta_n}(A_{\mathbf{S}_n,k})} \Gcal_n(\wb) \geq \veps(\alpha(\mathbf{S},n)) \Bigr\} \Biggr) \\ 
&\leq \zeta_2 + 2M\mathbb{E} \left[\sum_{m=0}^{\lceil \frac{d}{\rho_n} \rceil }\re^{-\frac{n\veps^2(m\rho_n)}{2\nu^2}}\left[\frac{1}{\beta_n} \right]^{\tilde{\alpha}(\Sb,n,k,\epsilon)} \times \mathds{1}\left\{\alpha(\mathbf{S},n) \in J_m(\rho_n) \right\}\right]\\
&\leq \zeta_2 +  2M
\mathbb{E} \left[\sum_{m=0}^{\lceil \frac{d}{\rho_n} \rceil }\re^{-\frac{n\veps^2(\alpha(\mathbf{S},n))}{2\nu^2} + 1}\left[\frac{1}{\beta_n} \right]^{\tilde{\alpha}(\Sb,n,k,\epsilon)} \times \mathds{1}\left\{\alpha(\mathbf{S},n) \in J_m(\rho_n) \right\}\right]\\
&= \zeta_2 + 2M\mathbb{E} \left[\re^{-\frac{n\veps^2(\alpha(\mathbf{S},n))}{2\nu^2} + 1}\left[\frac{1}{\beta_n} \right]^{\tilde{\alpha}(\Sb,n,k,\epsilon)} \right].
\end{align*}
By the definition of $\veps(t)$, for any $\Sb$ and $n$ we have that:
\begin{align*}
2M\re^{-\frac{n\veps^2(\alpha(\mathbf{S},n))}{2\nu^2} + 1}\left[\frac{1}{\beta_n} \right]^{\tilde{\alpha}(\Sb,n,k,\epsilon)}  
&=2 \re \zeta_2 .
\end{align*} 
Therefore,   
\begin{align*}
\pr \Biggl( \{\Sb \in \mathfrak{Z}\}\cap \Bigl\{ \max_{\wb \in N_{\beta_n}(A_{\mathbf{S}_n,k})} \Gcal_n(\wb) \geq \veps(\alpha(\mathbf{S},n)) \Bigr\} \Biggr) \leq (1+2\re)\zeta_2. 
\end{align*}

Therefore, by using the definition of $\veps(t)$, \eqref{eqn:proof_intermbound}, and \eqref{eqn:interm_lip}, with probability at least $1- \zeta - \delta_k - (1+2\re)\zeta_2$, we have
\begin{align*}
|\hat{\Rcal}(W,\Sb_n) - \Rcal(W)| 
\leq 2\sqrt{\frac{2 \nu^2}{n} \left[\log\left(\frac1{\beta_n}\right)\Bigl(\alpha(\Sb,n) + \xi_n c(\delta_k) + \xi_n c'(\beta_n) + \epsilon \Bigr)+ \log\left(\frac{M}{\zeta_2}\right) \right] }  
\nonumber
\\
\qquad\qquad\qquad+ 2L\beta_n.  
\end{align*}
Choose $k$ such that $\delta_k \leq \zeta/2$, $\zeta_2 = \zeta/(2+4\re)$, $\xi_n = \log\log(n)$, $\epsilon = \alpha(\Sb,n)$, and $\beta_n = \sqrt{2\nu^2/L^2 n}$. Then, with probability at least $1- 2\zeta$, we have
\begin{align}
&|\hat{\Rcal}(W,\Sb_n) - \Rcal(W)| \\
&\leq 4\sqrt{\frac{4\nu^2}{n} \left[\frac1{2}\log\left(nL^2\right)\Bigl(2\alpha(\Sb,n) +  c(\delta_k) \log\log(n) + o(\log\log(n))  \Bigr)+ \log\left(\frac{13M}{\zeta}\right) \right] }.  
\end{align}

Finally, as we have $\alpha(\Sb,n) \log(n) = \omega(\log\log(n)) $, for $n$ large enough, we obtain
\begin{align}
|\hat{\Rcal}(W,\Sb_n) - \Rcal(W)| \leq 8\nu\sqrt{ \frac{ \alpha(\Sb,n) \log^2\left(nL^2\right)  } {n}+ \frac{\log\left({13M}/{\zeta}\right)}{n}  } . 
\end{align}
This completes the proof.
\end{proof}

\subsection{Proof of Proposition~\ref{cor:lse}}

\begin{proof}
If we apply SGD 
this results in the recursion \eqref{eq-ifs-2} with
\begin{align}\label{eq-stoc-grad-2}
&h_i(w) = M_i w  + q_{i} \quad\text{ with} \quad M_{i}:=(1-\eta\lambda) I - \eta {H_{i}}, 
\\
&H_i := \frac{1}{b}\sum\limits_{j\in S_i } a_j a_j^T, \quad q_i := (\eta/b) \sum\limits_{j\in S_i } a_j y_j\,,
\nonumber
\end{align}

where $a_j \in \mathbb{R}^d$ are the input vector, and $y_j$ are the output variable, 
and $\{S_{i}\}_{i=1}^{m_{b}}$ is a partition of $\{1,2,\dots,n\}$ with $|S_i| = b$ with $i=1,2,\ldots,m_{b}$ and $m_{b}=n/b$. Let $L_i$ be the Lipschitz constant of $a_i$. It can be seen that $\nabla \ell(w,z_{i})$ is Lipschitz with constant $L_i = R_i^2 + \lambda$, where $R_i = \max_{j\in S_i} \|a_j\|$.  We assume
$\eta < 2/L = 2/(R^2+ \lambda)$, where $R = \max_i R_i$, otherwise the expectation of the iterates can diverge from some initializations and for some choices of the batch-size. We have
$$h_i(u) - h_i(v) = M_i (u-v),$$
where   
$$ 0 \preceq \left(1- \eta\lambda - \eta R_i^2\right) I  \preceq M_i \preceq  (1- \eta \lambda) I.$$ 
Hence, $h_i$ is bi-Lipschitz in the sense of \cite{anckar2016dimension} where
$$\gamma_i (u-v)\leq\|h_i(u) - h_i(v)\| 
\leq \Gamma_i (u-v),$$
with 
\begin{align}
&\gamma_i = \min\left(\left|1- \eta\lambda - \eta R_i^2\right|, \left|1- \eta \lambda\right|\right),
\\
&\Gamma_i =  \max\left(\left|1- \eta\lambda - \eta R_i^2\right|, \left|1- \eta \lambda\right|\right)<1,
\end{align}
as long as $\gamma_i > 0.$ For simplicity of the presentation, we  assume 
$\eta < \frac{1}{R^2 + \lambda}$ in which case the expressions for $ \gamma_i $ and $ \Gamma_i $ simplify to: 
    $$\gamma_i = 1- \eta\lambda - \eta R_i^2, \quad \Gamma_i = 1- \eta \lambda.$$
In this case, it is easy to see that 
$$0 < \gamma_i \leq \|J_{h_i}(w)\| \leq \Gamma_i < 1,$$ and it follows from Theorem~\ref{thm-rams} that 
\begin{equation}\label{def-lower-upper-dim}
\dimhu\mu_{W|\Sb_n} \leq \frac{\mathrm{Ent}}{\sum_{i=1}^{m_{b}} p_i \log(\Gamma_i)}=\frac{-\mathrm{Ent}}{\sum_{i=1}^{m_{b}} p_i \log(1/\Gamma_i)}.
\end{equation}

By Jensen's inequality, we have
\begin{equation}\label{Ent:ineq}
-\mathrm{Ent}=\sum_{i=1}^{m_{b}}p_{i}\log\left(\frac{1}{p_{i}}\right)\leq\log\left(\sum_{i=1}^{m_{b}}p_{i}\cdot\frac{1}{p_{i}}\right)=\log(m_{b}),
\end{equation}
where $m_{b}=\binom{n}{b}$.

When $\eta<\frac{1}{R^{2}+\lambda}$, we recall that $\gamma_{i}=1-\eta\lambda-\eta R_{i}^{2}$
and $\Gamma_{i}=1-\eta\lambda$. Therefore,

\begin{equation}
\dimhu \mu_{W|\Sb_n} \leq
\frac{-\mathrm{Ent}}{\sum_{i=1}^{m_{b}} p_i \log(1/\Gamma_i)}
\leq\frac{\log\left(m_{b}\right)}{\log(1/(1-\eta\lambda))} 
=\frac{\log\left(n/b\right)}{\log(1/(1-\eta\lambda))}. 
\end{equation}

The proof is complete.
\end{proof}

\subsection{Proof of Proposition~\ref{cor:logistic}}

\begin{proof}
When the batch-size is equal to $b$, we can compute that the Jacobian is given by
\begin{equation} 
    J_{h_i}(w) = \frac{1}{b}\sum_{j\in S_{i}}\left(1 - \eta \lambda + \eta y_j^2 \left[ \frac{e^{-y_j a_j^T w}}{(1 + e^{-y_j a_j^T w})^2} 
    \right]
    a_j a_j^T\right),
\end{equation}
where $\{S_{i}\}_{i=1}^{m_{b}}$ is a partition of $\{1,2,\dots,n\}$ with $|S_i| = b$, where $i=1,2,\ldots,m_{b}$ and $m_{b}=n/b$.
Note that the input data is bounded, i.e. $R_{i}:=\max_{j\in S_{i}}\|a_j\|<\infty$, 
and $R:=\max_{i}R_{i}<2\sqrt{\lambda}$.
Recall that the step-size is sufficiently small, i.e. $\eta<1/\lambda$. 
One can provide the upper bound 
on $J_{h_i}(w)$:
\begin{align}\label{uses:ineq}
\Vert J_{h_{i}}(w)\Vert
\leq\Gamma_{i}:=1-\eta\lambda+\frac{1}{4}\eta R_{i}^{2}\leq 1-\eta\lambda+\frac{1}{4}\eta R^{2},
\end{align}
so that
\begin{align}
\dimhu\mu_{W|\Sb_n}
&\leq
\frac{\mathrm{Ent}}{\sum_{i=1}^{m_{b}} p_i \log(\Gamma_i)}
=\frac{-\mathrm{Ent}}{\sum_{i=1}^{m_{b}}p_{i}\log(1/(1-\eta\lambda+\frac{1}{4}\eta R_{i}^{2}))}
\nonumber
\\
&\leq
\frac{\log m_{b}}{\log(1/(1-\eta\lambda+\frac{1}{4}\eta R^{2}))}\label{second:last:ineq}
\\
&=\frac{\log\left(n/b\right)}{\log(1/(1-\eta\lambda+\frac{1}{4}\eta R^{2}))},\label{last:ineq}
\end{align}
where we used \eqref{uses:ineq} and \eqref{Ent:ineq} in \eqref{second:last:ineq}.

The proof is complete.
\end{proof}

\subsection{Proof of Proposition~\ref{cor:nonconvex:logistic}}

\begin{proof}
We can compute that 
\begin{align}
&\nabla \ell(w,z_{i}) = -a_i \rho'\left(y_i - \langle w,a_i \rangle \right) + \lambda_r w,
\\
&h_i(w) = \frac{1}{b}\sum_{j\in S_{i}}(1- \eta \lambda_r)w + \eta a_j \rho'\left(y_j - \langle w,a_j \rangle \right),
\\
&J_{h_i}(w) =\frac{1}{b}\sum_{j\in S_{i}} (1- \eta \lambda_r)I - \eta a_j a_j^T          \rho''\left(y_j - \langle w,a_j \rangle \right),   
\end{align}
where $\{S_{i}\}_{i=1}^{m_{b}}$ is a partition of $\{1,2,\dots,n\}$ with $|S_i| = b$, where $i=1,2,\ldots,m_{b}$ with $m_{b}=n/b$.
Furthermore, $\|\rho_{\exp}''\|_\infty = \rho_{\exp}''(0) = \frac{2}{t_0}$. Therefore, for $\eta \in (0, \frac{1}{\lambda_r + R^2(2/t_0)})$,
\begin{equation}\label{uses:ineq:2}
0<(1- \eta \lambda_r) - \eta R^2 \frac{2}{t_0} \leq \|J_{h_i}(w)\| \leq  (1- \eta \lambda_r) + \eta R^2 \frac{2}{t_0},
\end{equation}
where $R = \max_i \|a_i\|<\sqrt{\lambda_{r}t_{0}/2}$. We have
\begin{equation}
\dimhu\mu_{W|\Sb_n}\leq\frac{\log m_{b}}{\log(1/(1- \eta \lambda_r + \eta R^2 \frac{2}{t_0}))}
=\frac{\log\left(n/b\right)}{\log(1/(1- \eta \lambda_r + \eta R^2 \frac{2}{t_0}))},
\end{equation}
where we used \eqref{uses:ineq:2} and \eqref{Ent:ineq}.
The proof is complete.

\end{proof}

\subsection{Proof of Proposition~\ref{ex-svm}}

\begin{proof}
We can compute that
\begin{align*}
&\nabla \ell(w,z_{i}) = y_i \ell_\sigma'\left(y_i a_i^T w\right) a_i + \lambda w,
\\
&\nabla^2 \ell(w,z_{i}) = y_i^2 \ell_\sigma''\left(y_i a_i^T w\right) a_i a_i^T + \lambda,
\\
&h_i(w) = w - \frac{\eta}{b}\sum_{j\in S_{i}} \nabla \ell(w,z_{j}),
\end{align*}
where $\{S_{i}\}_{i=1}^{m_{b}}$ is a partition of $\{1,2,\dots,n\}$ with $|S_i| = b$, where $i=1,2,\ldots,m_{b}$ with $m_{b}=n/b$,
so that 
    $$ 
    J_{h_i}(w) = I - \frac{\eta}{b} \sum_{j\in S_{i}}\nabla^2 \ell(w,z_{j}) 
    = (1- \eta \lambda)I - \frac{\eta}{b}\sum_{j\in S_{i}} y_j^2 \ell_\sigma''\left(y_j a_j^T w\right) a_j a_j^T,
    $$
with    

    $$ \ell_\sigma''(z) = \frac{1}{\sigma} \frac{e^{-(1-z)/\sigma}}{(1+e^{-(1-z)/\sigma})^2} \geq 0, \quad \|\ell_\sigma''\|_\infty =\ell_\sigma''(1) =  \frac{1}{4\rho}.$$
Therefore, if $\eta \in (0, \frac{1}{\lambda + \|R\|^2/(4\rho)})$ and $R := \max_i \|a_i\|$, then
\begin{equation*}
1-\eta \lambda-\eta \frac{1}{4\rho}R^2   \leq \|J_{h_i}(w)\| \leq 1-\eta \lambda.
\end{equation*}
This implies that
\begin{equation}
\dimhu\mu_{W|\Sb_n}\leq\frac{\log m_{b}}{\log(1/(1-\eta\lambda))}=\frac{\log\left(n/b\right)}{\log(1/(1-\eta\lambda))},
\end{equation}
where we used \eqref{Ent:ineq}.
The proof is complete.
\end{proof}

\subsection{Proof of Proposition~\ref{ex:one:hidden:layer}}

\begin{proof}
We recall that the loss is given by:
\begin{equation}\label{eqn:one:hidden:appendix}
    \ell(w,z_{i}):=\|y_i -\hat{y}_i\|^2 + \lambda\|w\|^2/2, \quad  \hat{y}_i := \sum\limits_{r=1}^m b_r \sigma\left( w_r^T a_i\right),  
\end{equation}
where the non-linearity $\sigma:\mathbb{R}\to\mathbb{R}$ is smooth and $\lambda>0$ is a regularization parameter. 
Note that we can re-write \eqref{eqn:one:hidden:appendix} as 
$\ell(w,z_{i})= \left\|y_i -  b^T \sigma\left( w_r^T a_i\right) \right\|^2 + \lambda \|w\|^2/2$.
We can compute that
\begin{equation} 
\frac{\partial \ell(w,z_{i})}{\partial w_r} = -(y_i - \hat{y}_i) \frac{\partial \hat{y}_i}{\partial w_r} +\lambda w_r =-(y_i - \hat{y}_i)b_r \sigma'(w_r^T a_i) a_i + \lambda w_r.
\end{equation}
Therefore, 
$$\nabla\ell(w,z_{i}) = -(y_i - \hat{y}_i) v_i  +\lambda w, \quad\text{where}\quad v_i:=\begin{bmatrix} 
b_1 \sigma'(w_1^T a_i)a_i \\
b_2 \sigma'(w_2^T a_i)a_i  \\
\cdots \\
b_m \sigma'(w_m^T a_i)a_i\\
\end{bmatrix}, 
$$ 
with
 $$ h_i(w) = w - \frac{\eta}{b}\sum_{j\in S_{i}}\nabla \ell(w,z_{i}), 
$$
and 
\begin{align}
J_{h_i}(w) 
 &= (1-\eta \lambda)I - \frac{\eta}{b}\sum_{j\in S_{i}} v_j \otimes v_j^T 
 \nonumber
 \\
 &\quad
 + \frac{\eta}{b}\sum_{j\in S_{i}} (y_j - \hat{y}_j) \begin{bmatrix} 
b_1 \sigma''(w_1^T a_j) a_ja_j^T &  0_d & \hdots & 0_d\\ 
0_d                               & b_2 \sigma'' (w_2^T a_j)a_ja_j^T &  \hdots & 0_d \\
\vdots & \vdots & \ddots & \vdots   \\
0_d & 0_d& \hdots & b_m \sigma''(w_m^T a_j)a_j a_j^T\\
\end{bmatrix}
\nonumber
\\
&= (1-\eta \lambda)I -\frac{\eta}{b}\sum_{j\in S_{i}} \mbox{diag}(\{B_r^{(j)}\}_{r=1}^m),  
\end{align} 
where $\{S_{i}\}_{i=1}^{m_{b}}$ is a partition of $\{1,2,\dots,n\}$ with $|S_i| = b$, where $i=1,2,\ldots,m_{b}$ with $m_{b}=n/b$,
and
\begin{equation}
B_r^{(i)} := b_r \left[-(y_i - \hat{y}_i) \sigma''(w_r^T a_i) +(\sigma'(w_r^T a_i))^2 \right] a_ia_i^T,
\end{equation} 
and $0_d$ is a $d\times d$ zero matrix and $\mbox{diag}(\{B_r^{(i)}\}_{r=1}^m)$ denotes a block diagonal matrix with the matrices $B_r^{(i)}$ on the diagonal. We assume the output $y_i$ and the activation function $\sigma$ and its second derivative $\sigma''$ is bounded. This would for instance clearly hold for classification problems where $y_i$ can take integer values on a compact set with a sigmoid or hyperbolic tangent activation function. Then, under this assumption, there exists a constant $M_{y}>0$ such that 
$\max_i \|y_i - \hat{y}_i \| \leq M_{y}$. Then for $\eta \in (0,  \frac{1}{2\lambda})$ and $\lambda > C$  where $C:=M_{y}\|b\|_\infty \|\sigma''\| R^2 +\left( \max_j \|v_j\|_\infty \right)^2$, we get
$$
1 - \eta (C + \lambda) \leq \|J_{h_i}(w) \| \leq 1 - \eta (\lambda - C).
$$
This implies that
\begin{equation}
\dimhu\mu_{W|\Sb_n}\leq\frac{\log m_{b}}{\log(1/(1-\eta(\lambda-C)))}=\frac{\log\left(n/b\right)}{\log(1/(1-\eta(\lambda-C)))},
\end{equation}
where we used \eqref{Ent:ineq}.
The proof is complete.

 \end{proof}

\subsection{Proof of Proposition~\ref{cor:least:squares:2}}

\begin{proof}
Recall that $H$ is positive-definite and
there exist some $m,M>0$:
\begin{equation}
0\prec mI\preceq H\preceq MI.
\end{equation}
We have
$$h_i(u) - h_i(v) = M_i (u-v),$$
where   
\begin{equation}
0 \preceq \left(1- \eta\lambda m^{-1} - \eta m^{-1} R_i^2\right) I  \preceq M_i \preceq  \left(1- \eta \lambda M^{-1}\right) I,
\end{equation}
where $R_{i}:=\max_{j\in S_{i}}\Vert a_j\Vert$, 
and we recall the assumption that
$\eta < \frac{m}{R^2 + \lambda}$, with $R:=\max_{i}R_{i}$.
Hence, $h_i$ is bi-Lipschitz in the sense of \cite{anckar2016dimension} where
$$\gamma_i (u-v)\leq\|h_i(u) - h_i(v)\| 
\leq \Gamma_i (u-v),$$
with 
\begin{align}
&\gamma_i = \min\left(\left|1- \eta\lambda m^{-1} - \eta m^{-1} R_i^2\right|, \left|1- \eta M^{-1}\lambda\right|\right),
\\
&\Gamma_i =  \max\left(\left|1- \eta\lambda m^{-1} - \eta m^{-1} R_i^2\right|, \left|1- \eta M^{-1}\lambda\right|\right)<1,
\end{align}
as long as $\gamma_i > 0.$ We recall the assumption 
$\eta < \frac{m}{R^2 + \lambda}$, where $R:=\max_{i}R_{i}$, in which case the expressions for $ \gamma_i $ and $ \Gamma_i $ simplify to: 
    $$\gamma_i = 1- \eta m^{-1}\lambda - \eta m^{-1} R_i^2, \quad \Gamma_i = 1- \eta M^{-1}\lambda.$$
In this case, it is easy to see that 
$$0 < \gamma_i \leq \|J_{h_i}(w)\| \leq \Gamma_i < 1,$$ and it follows from Theorem~\ref{thm-rams} that 
\begin{equation}
\dimhu\mu_{W|\Sb_n}\leq\frac{\mathrm{Ent}}{\sum_{i=1}^{m_{b}} p_i \log(\Gamma_i)}\leq\frac{\log m_{b}}{\log(1/(1-\eta M^{-1}\lambda))}
=\frac{\log\left(n/b\right)}{\log(1/(1-\eta M^{-1}\lambda))},
\end{equation}
where we used \eqref{Ent:ineq}.
The proof is complete.

\end{proof}

\subsection{Proof of Proposition~\ref{cor:logistic:2}}

\begin{proof}
Similar as in the proof of Proposition~\ref{cor:logistic},
we can compute that the Jacobian is given by
\begin{equation} 
    J_{h_i}(w) = \frac{1}{b}\sum_{j\in S_{i}}\left(1 - \eta H^{-1}\lambda + \eta H^{-1} y_j^2 \left[ \frac{e^{-y_j a_j^T w}}{(1 + e^{-y_j a_j^T w})^2} 
    \right]
    a_j a_j^T\right),
\end{equation}
where $\{S_{i}\}_{i=1}^{m_{b}}$ is a partition of $\{1,2,\dots,n\}$ with $|S_i| = b$, where $i=1,2,\ldots,m_{b}$ with $m_{b}=n/b$, and $H$ is a positive-definite matrix with
$0\prec mI\preceq H\preceq MI$.
recall that the input data is bounded, i.e. $\max_{j\in S_{i}}\|a_j\|\leq R_{i}$ for some $R_{i}$, 
and $R:=\max_{i}R_{i}$ satisfying $R<2\sqrt{m\lambda/M}$.
Also recall the step-size is sufficiently small, i.e. $\eta<m/\lambda$. 
One can provide upper bounds and lower bounds on $J_{h_i}(w)$:
\begin{align}
&\Vert J_{h_{i}}(w)\Vert
\leq\Gamma_{i}:=1-\eta M^{-1}\lambda+\frac{1}{4}\eta m^{-1}R_{i}^{2},
\\
&\Vert J_{h_{i}}(w)\Vert
\geq\gamma_{i}:=1-\eta m^{-1}\lambda,
\end{align}
so that

\begin{align}
\dimhu\mu_{W|\Sb_n}
&\leq\frac{-\mathrm{Ent}}{\sum_{i=1}^{m_{b}}p_{i}\log(1/(1-\eta M^{-1}\lambda+\frac{1}{4}\eta m^{-1}R_{i}^{2}))}
\nonumber
\\
&\leq\frac{\log m_{b}}{\log(1/(1-\eta M^{-1}\lambda+\frac{1}{4}\eta m^{-1}R^{2}))}\label{so:that:1}
\\
&=\frac{b\log\left(n/b\right)}{\log(1/(1-\eta M^{-1}\lambda+\frac{1}{4}\eta m^{-1}R^{2}))},\label{so:that:2}
\end{align}
where we used \eqref{Ent:ineq} in \eqref{so:that:1}.
The proof is complete.

\end{proof}

\subsection{Proof of Proposition~\ref{cor:nonconvex:logistic:2}}

\begin{proof}
Similar as in the proof of Proposition~\ref{cor:nonconvex:logistic},
we can compute that 
\begin{align}
J_{h_i}(w) =\frac{1}{b}\sum_{j\in S_{i}} \left(I- \eta H^{-1}\lambda_r\right) - \eta H^{-1}a_j a_j^T          \rho''\left(y_j - \langle w,a_j \rangle \right), 
\end{align}
where $\{S_{i}\}_{i=1}^{m_{b}}$ is a partition of $\{1,2,\dots,n\}$ with $|S_i| = b$, where $i=1,2,\ldots,m_{b}$ with $m_{b}=n/b$,
and $H$ is a positive-definite matrix with $0\prec mI\preceq H\preceq MI$.
For the function $\rho$, a standard choice 
is exponential squared loss: $\rho_{\exp}(t) = 1 - e^{-|t|^2/t_0}$, where $t_0>0$ is a tuning parameter.
We can compute that $\|\rho_{\exp}''\|_\infty = \rho_{\exp}''(0) = \frac{2}{t_0}$. Therefore, for $\eta \in (0, \frac{m}{\lambda_r + R^2(2/t_0)})$,
\begin{equation} 
0< 1- \eta m^{-1}\lambda_r - \eta m^{-1} R^2 \frac{2}{t_0} \leq \|J_{h_i}(w)\| \leq  1- \eta M^{-1}\lambda_r + \eta m^{-1}R^2 \frac{2}{t_0},
\end{equation}
where $R = \max_i \|a_i\|$ and we recall that $R<\sqrt{\lambda_{r}t_{0}m/(2M)}$. We have
\begin{equation}
\dimhu\mu_{W|\Sb_n}
\leq\frac{\log m_{b}}{\log(1/(1- \eta M^{-1} \lambda_r + \eta m^{-1}R^2 \frac{2}{t_0}))}
=\frac{\log\left(n/b\right)}{\log(1/(1- \eta M^{-1} \lambda_r + \eta m^{-1}R^2 \frac{2}{t_0}))},
\end{equation}
where we used \eqref{Ent:ineq}.
The proof is complete.

\end{proof}

\subsection{Proof of Proposition~\ref{cor:svm:2}}

\begin{proof}
Similar as in the proof of Proposition~\ref{ex-svm}, we can compute that
\begin{equation*} 
    J_{h_i}(w) = I - \frac{\eta}{b} H^{-1}\sum_{j\in S_{i}}\nabla^2 \ell(w,z_{j})
    = (1- \eta \lambda H^{-1})I - \frac{\eta}{b}H^{-1}\sum_{j\in S_{i}} y_j^2 \ell_\sigma''\left(y_j a_j^T w\right) a_j a_j^T,
\end{equation*}
where $\{S_{i}\}_{i=1}^{m_{b}}$ is a partition of $\{1,2,\dots,n\}$ with $|S_i| = b$, where $i=1,2,\ldots,m_{b}$ with $m_{b}=n/b$,
and $H$ is a positive-definite matrix with
$0\prec mI\preceq H\preceq MI$,
and
$$ \ell_\sigma''(z) = \frac{1}{\sigma} \frac{e^{-(1-z)/\sigma}}{(1+e^{-(1-z)/\sigma})^2} \geq 0, \quad \|\ell_\sigma''\|_\infty =\ell_\sigma''(1) =  \frac{1}{4\rho}.$$
Therefore, if $\eta \in (0, \frac{m}{\lambda + \|R\|^2/(4\rho)})$ where $R := \max_i \|a_i\|$, then
$$ 1-\eta m^{-1}\lambda-\eta m^{-1}\frac{1}{4\rho}R^2   \leq \|J_{h_i}(w)\| \leq 1-\eta M^{-1}\lambda. $$
This implies that
\begin{equation}
\dimhu\mu_{W|\Sb_n}\leq\frac{\log m_{b}}{\log(1/(1-\eta M^{-1}\lambda))}
=\frac{\log\left(n/b\right)}{\log(1/(1-\eta M^{-1}\lambda))},
\end{equation}
where we use \eqref{Ent:ineq}.
The proof is complete.

\end{proof}

\subsection{Proof of Proposition~\ref{cor:one:hidden:2}} 
 
\begin{proof} 
By following the similar derivations as in Proposition~\ref{ex:one:hidden:layer}, we obtain
\begin{align}
 J_{h_i}(w) 
 = \left(1-\eta \lambda H^{-1}\right)I -\frac{\eta}{b}H^{-1}\sum_{j\in S_{i}} \mbox{diag}\left(\{B_r^{(j)}\}_{r=1}^m\right),  
\end{align} 
where $\{S_{i}\}_{i=1}^{m_{b}}$ is a partition of $\{1,2,\dots,n\}$ with $|S_i| = b$, where $i=1,2,\ldots,m_{b}$, with $m_{b}=n/b$,
and $H$ is a positive-definite matrix and
$0\prec mI\preceq H\preceq MI$ 
for some $m,M>0$,
and $\mbox{diag}(\{B_r^{(i)}\}_{r=1}^m)$ denotes a block diagonal matrix with the matrices $B_r^{(i)}$ on the diagonal
defined in Proposition~\ref{ex:one:hidden:layer}. As in Proposition~\ref{ex:one:hidden:layer}, there exists a constant $M_{y}>0$ such that 
$\max_i \|y_i - \hat{y}_i \| \leq M_{y}$. Then for $\eta \in (0,  \frac{m}{C+\lambda})$ and $\lambda > \frac{M}{m}C$  where $C:=M_{y}\|b\|_\infty \|\sigma''\| R^2 +\left( \max_j \|v_j\|_\infty \right)^2$, we get
$$
1 - \eta m^{-1}(C + \lambda) \leq \|J_{h_i}(w) \| \leq 1 - \eta \left(M^{-1}\lambda-m^{-1}C\right).
$$
This implies that
\begin{equation}
\dimhu\mu_{W|\Sb_n}\leq\frac{\log m_{b}}{\log(1/(1-\eta(M^{-1}\lambda-m^{-1}C)))}
=\frac{\log\left(n/b\right)}{\log(1/(1-\eta(M^{-1}\lambda-m^{-1}C)))},
\end{equation}
where we used \eqref{Ent:ineq}.
The proof is complete.
\end{proof}

\end{document}